\documentclass[runningheads]{llncs}

\usepackage{eccv}

\usepackage{eccvabbrv}

\usepackage{graphicx}
\usepackage{booktabs}
\usepackage{subcaption}
\usepackage{multirow}
\usepackage{array}
\usepackage[dvipsnames, table]{xcolor}
\usepackage{pifont}
\usepackage{amssymb}
\usepackage{arydshln}
\newcommand{\cmark}{\ding{51}}%
\newcommand{\xmark}{\ding{55}}%
\usepackage{algorithm,algorithmic}
\usepackage{placeins}
\usepackage{diagbox}

\usepackage[accsupp]{axessibility}  % Improves PDF readability for those with disabilities.

\usepackage{hyperref}

\usepackage{orcidlink}

\definecolor{mplBlue}{RGB}{31,119,180}
\definecolor{mplOrange}{RGB}{255,127,15}

\renewcommand{\algorithmicrequire}{{\color{Fuchsia}\textbf{Require:}}}
\renewcommand{\algorithmicinput}{{\color{Fuchsia}\textbf{Input:}}}
\renewcommand{\algorithmicoutput}{{\color{Fuchsia}\textbf{Output:}}}
\renewcommand{\algorithmicensure}{{\color{Fuchsia}\textbf{Return:}}}

\renewcommand{\algorithmicfor}{{\color{Violet}\textbf{for}}}
\renewcommand{\algorithmicif}{{\color{Violet}\textbf{if}}}
\renewcommand{\algorithmicthen}{{\color{Violet}\textbf{then}}}
\renewcommand{\algorithmicdo}{{\color{Violet}\textbf{do}}}
\renewcommand{\algorithmicend}{{\color{Violet}\textbf{end}}}
\renewcommand{\algorithmiccomment}[1]{~{\color{CadetBlue}// #1}}

\begin{document}

\title{TASSO: TAsk-Specific Subspace Optimization for Continual Learning of Vision-Language Models} 

\titlerunning{TASSO: TAsk-Specific Subspace Optimization}

\author{Chang Sun\orcidlink{0009-0004-5086-8286} \and
Francesco Barbato\orcidlink{0000-0001-9893-5813} \and
Matteo Caligiuri\orcidlink{0009-0006-2928-1047} \and
Pietro Zanuttigh\orcidlink{0000-0002-9502-2389}}

\authorrunning{C.~Sun et al.}

\institute{University of Padova, Padova(PD), Italy \\
% \and
% Springer Heidelberg, Tiergartenstr.~17, 69121 Heidelberg, Germany
% \url{http://www.springer.com/gp/computer-science/lncs} 
% \and
\email{\{sunchang,francesco.barbato,matteo.caligiuri,zanuttigh\}@dei.unipd.it}}

\maketitle

\begin{abstract}
Vision-Language Models (VLMs) exhibit strong zero-shot capabilities, making them an attractive solution for continual learning across diverse tasks. However, during continual adaptation, both catastrophic forgetting and zero-shot degradation occur, severely degrading performance. 
In this paper, we introduce TASSO\footnote{Code is available at \url{https://github.com/LTTM/TASSO}}, a new paradigm that efficiently preserves the latent space geometry while ensuring network plasticity. We achieve this with two complementary techniques: subspace learning and geometry-aware knowledge distillation.
Specifically, we first learn a sequence of task-specific low-rank projectors, which we use to project the latent representations before optimizing cross-entropy. Secondly, we employ a geodesic-distance-based loss that distills knowledge from the previous-task model while effectively preserving the latent space geometry. 
These design choices not only avoid unnecessary parameter updates along the full embedding dimensions but also improve learning by focusing on task-specific manifolds.
Moreover, the geometry-aware distillation provides strong regularization and significantly reduces both catastrophic forgetting and zero-shot degradation throughout the continual learning sequence. 
Experimental results with the CLIP vision language model in the multi-domain task incremental and class incremental learning benchmarks demonstrate clear improvements over state-of-the-art methods in mitigating forgetting and preserving zero-shot capabilities.

\end{abstract}

\section{Introduction}
\label{sec:intro}
Recently, highly performing Vision-language Models (VLMs), such as CLIP~\cite{radford2021learning}, BLIP~\cite{li2022blip}, ALIGN~\cite {jia2021scaling}, and Flamingo~\cite{alayrac2022flamingo}, have emerged as key tools for Artificial Intelligence (AI), bridging Computer Vision (CV) and Natural Language Processing (NLP), and enabling machines to jointly understand images and text, thus producing valuable information for many downstream tasks. 

While their zero-shot accuracy is undeniable, there exist several deployment scenarios, \eg, embodied agents, autonomous driving, or robotics, where static general-level knowledge may not be sufficient. To effectively tackle these issues, VLMs must be able to continuously learn new and specific concepts, refining their internal knowledge over time~\cite{liu2025continual}.

The Continual Learning (CL) task has been thoroughly investigated across various computer vision tasks and architectures~\cite{wang2024comprehensive}, but the specific characteristics of VLMs introduce several additional issues that require attention \cite{liu2025continual}.

More specifically, in addition to the standard issue of \textit{Catastrophic Forgetting} in continual learning (\ie, optimizing models for new tasks leads to degraded performance on previous ones), the Open-Vocabulary nature of VLMs also necessitates addressing \textit{Zero-Shot Degradation} (\ie, when learning a task degrades accuracy on future, unrelated tasks).

Tackling these complementary issues requires a careful balance between achieving accuracy in the new task being learned, preservation of old knowledge, and maintenance of zero-shot capability, as the \textit{plasticity-stability tradeoff} is even more challenging in this setting.

Current methods tackle the problem using various distillation strategies from the pre-trained model, which reduce forgetting but typically also lead to a diminished capability to learn new tasks. Other approaches add external adapters with additional parameters, allowing for the learning of new tasks at the expense of cumbersome architectures that do not scale well to a large number of tasks.

To this end, we introduce a novel continual learning strategy for VLMs, which we denote TASSO (TAsk-Specific Subspace Optimization for Continual Learning of Vision-Language Models). 
It introduces a subspace learning strategy that separates knowledge related to the current task from general knowledge, along with a novel distillation objective that exploits the geodesic distance. 

The subspace separation allows TASSO to optimize the training process of the new task by focusing only on the relevant manifold and avoiding unnecessary parameter updates that may impair performance on previous or future tasks. Moreover, we leverage these orthogonal subspaces to distill knowledge from prior tasks more effectively: we explicitly extract task-relevant information and enable the training process to separate it from newly learned content, while task-irrelevant information can remain fully preserved. 

The experimental evaluation under the Multidomain Class Incremental and  Task Incremental Learning (MCIL and MTIL) settings demonstrates how these provisions enable us to achieve state-of-the-art performance on challenging continual learning benchmarks for VLMs. 
In summary, our contributions are:
\begin{enumerate}
    \item We propose a novel framework that efficiently tackles catastrophic forgetting and zero-shot degradation through low-rank subspace learning.
    \item We introduce a learned projector that separates the subspace relevant to the current task from the irrelevant part and selectively applies the losses.
    \item We present a novel geometry- and subspace-aware loss for knowledge distillation based on the geodesic distance.
    \item Results show how the approach outperforms the state-of-the-art, demonstrating an impressive capability to tackle zero-shot degradation and catastrophic forgetting. 
\end{enumerate}

%%%%%%%%%%%%%%%%%%%%%%%%%%%%%%%%%%%%%%%%%%%%%%%%%%%%%%%%%%%%%%%%%%%%%%%%%%%%%%%%%%%%%%%%%%%%%%%%%%
\section{Related Works}
\label{sec:relatedworks}
\subsubsection{Continual Learning for Vision-Language Models}
Recent continual learning work has increasingly focused on pretrained vision-language models (especially CLIP), where the goal is to adapt to sequential tasks while preserving both previously learned knowledge and the pretrained model's generalization ability. 

Early CLIP-based methods, such as \cite{wang2022s}, utilize prompt tuning with a frozen backbone, demonstrating that adaptation methods can perform well in domain-incremental settings. A key VLM-specific challenge was later highlighted by the authors of \cite{zheng2023preventing}, who introduced the MTIL benchmark and discovered that continual learning can also hinder zero-shot transfer. 
More recent work broadens the setting to open-domain continual learning \cite{li2025coleclip}, multimodal image-caption streams \cite{liu2025c}, and time-continual CLIP training at web scale \cite{garg2023tic}.

\subsubsection{Distillation-based Continual Adaptation}
The key objective of continual learning in VLMs is to mitigate both catastrophic forgetting and zero-shot degradation. Distillation-based methods tackle this problem by transferring knowledge from previous models to the model being trained~\cite{li2024continual}. 

More specifically, \cite{zheng2023preventing} proposed distilling knowledge from the pretrained CLIP using reference images from a public dataset, while preserving its zero-shot capabilities. 
Furthermore, \cite{yu2024select, zheng2024adapt} proposed utilizing two teachers in parallel: the pretrained model and the model from the previous task. 
The approach in \cite{yu2024select} focuses on weighting the two teachers differently based on each reference image. The method of \cite{zheng2024adapt} instead aims to distill multimodal proximity while preserving the intra- and inter-modal information from both the vision and text modalities. 

These methods all rely on the distillation of the pretrained model for zero-shot degradation mitigation, which significantly increases storage and computational costs and may risk damaging the ability to learn new tasks.

\subsubsection{Parameter-Efficient Adaptation}
An alternative to KD-based approaches is parameter-efficient training. This technique optimizes a limited set of parameters to rapidly adapt a pretrained model to downstream tasks. Popular approaches include Low-Rank Approximation (LoRA)~\cite{hu2022lora} and adapter modules~\cite{houlsby2019parameter, wang2021k}. 

Continual low-rank learning is studied in \cite{lu2024adaptive}, which systematically analyzes how the rank and placement of LoRA modules can affect learning and forgetting. 
Their findings confirm the intuition that a relatively high-rank LoRA improves task learning but increases forgetting, while a relatively low-rank LoRA reduces forgetting but limits adaptation. Their proposed solution involves adaptively changing the LoRA rank to achieve an optimal plasticity-stability balance. 

An interesting refinement is introduced by \cite{yu2024boosting}, which exploits the idea of Mixture-of-Experts (MoE)~\cite{jacobs1991adaptive} for the continual learning of CLIP. Their method dynamically expands the pretrained CLIP model through the integration of MoE adapters in response to new tasks. To preserve the zero-shot ability, they also designed an Auto-Selector that can automatically route inputs into the MoE adapters or the pretrained CLIP. 

A strategy that handles the two modalities in the CLIP model asymmetrically is proposed in \cite{kang2025dynamic}. It restricts the update of visual parameters within the common subspace of multiple null spaces, further limiting the impacts of non-zero residual terms. Despite their efficiency, existing methods rarely address explicit plasticity restoration, and they do not leverage task-specific structure in the feature subspace.

Our method balances parameter-efficient strategies with knowledge distillation by transferring knowledge from prior models through geometry-aware metrics, while constraining the optimization of the VLM on the current task to a low-rank subspace, all without explicitly adding adapter modules.

%%%%%%%%%%%%%%%%%%%%%%%%%%%%%%%%%%%%%%%%%%%%%%%%%%%%%%%%%%%%%%%%%%%%%%%%%%%%%%%%%%%%%%%%%%%%%%%%%%%%
\section{Continual Learning Setup}
\label{sec:preliminary}
In this section, we introduce the mathematical notation for the Task and Class-Incremental Learning~\cite{van2019three} settings used throughout the paper.
The objective is the continual optimization of a Vision Language Model $g$ (in our experiments, we used CLIP \cite{radford2021learning}) across a sequence of $K$ tasks $\mathcal{T} = \{\mathcal{T}^1, \cdots, \mathcal{T}^K\}$. Each task $T$ is associated with a set of $N$ labeled images, $\mathcal{D}$, and the corresponding class names $\mathcal{C}$.
In task $T^k = \{(\mathcal{D}^k, \mathcal{C}^k)\}$, the  model $g^k$ is trained on samples $\{(\mathbf{x}^k_j, \mathbf{y}^k_j)\} \in \mathcal{D}^k$, where $x^k_j$ is the j-th image in the dataset and $\mathbf{y}^k_j$ is the corresponding one-hot encoded label. During training, the label $\mathbf{y}^k_j$ is also used to extract the appropriate class name $c^k_j$ from $\mathcal{C}$ for use in the textual prompt~\cite{radford2021learning}.
Each class $c$ in task $\mathcal{C}^k$ is transformed using the template ``The photo of \{c\}'', before being encoded by the textual branch of the VLM $g_t$ into the $\mathbf{f}_t$ embedding. Similarly, an image $x$ is fed to the vision branch of CLIP $g_i$ to obtain the visual embedding $\mathbf{f}_i$. Note that the feature vectors are normalized before use, \ie, $\left|\left| \mathbf{f}_i \right|\right|_2 = 1$, $\left|\left| \mathbf{f}_t \right|\right|_2 = 1$.
Following the standard approach, to perform classification, the vision and textual embeddings are compared using cosine similarity (denoted as $\left< \cdot, \cdot \right>$). More specifically, a given vision embedding is compared to the $\mathbf{f}_t$'s of all classes in $\mathcal{C}$, and the closest vector is taken as the prediction: $\hat{y} = \text{argmax}_{c\in\mathcal{C}} \left<\mathbf{f}_i, g_t(c)\right>$.
In multi-task settings, inference can be performed in a task-specific or task-agnostic manner~\cite{yu2024select}. The former setting, denoted as MTIL (Multidomain Task Incremental Learning), assumes that each test sample is accompanied by a task index $k$ that allows for the selection of the appropriate class set $\mathcal{C}^k$ for classification. The latter, denoted as MCIL (Multidomain Class Incremental Learning), removes this assumption and makes predictions on the full label set given by concatenating all the $\mathcal{C}^k$.

\begin{figure}[t]
    \centering
    \includegraphics[width=\linewidth]{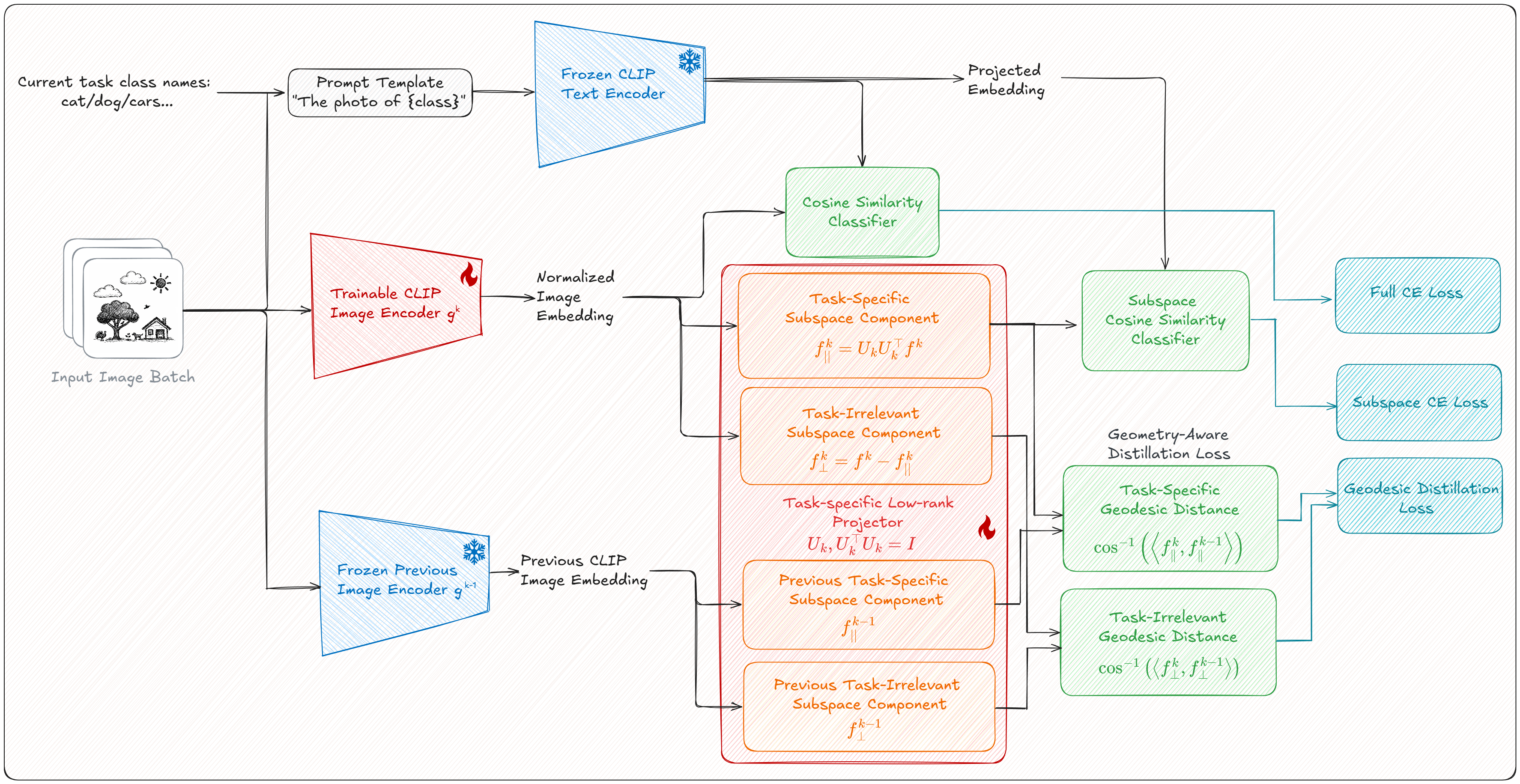}
    \caption{TASSO architecture. At each incremental step $k$, the vision encoder $g^k_i$ is optimized with a combination of three objectives: cross-entropy; task-specific subspace cross-entropy (computed according to the learnable projector $U$); and geometry-aware knowledge distillation (retaining knowledge from the previous step's encoder $g_i^{k-1}$).}
    \label{fig:arch}
\end{figure}

\section{Continual Learning Strategy}
\label{sec:strategy}
To perform continual optimization of the VLM, we start from using a standard Cross-Entropy (CE) objective computed on the distribution of the similarities of the vision-textual embeddings. More specifically, the core optimization target is:
\begin{equation}
    \mathcal{L}_{CE} = \frac{1}{N}\sum_{j=1}^N \text{CE}(\mathbf{s}_j, \mathbf{y}_j)\;,
\end{equation}
where $\mathbf{s} = \text{softmax}([\left<\mathbf{f}_i, g_t(c)\right> \; \forall c \in \mathcal{C}])$ is the vector of class probabilities for a sample $x$. Note that we only train the vision branch; the text encoder is frozen.

When cross-entropy is used as the sole optimization objective, training VLMs over a series of incremental tasks leads to two undesirable effects: \textit{Catastrophic Forgetting} of previous tasks' knowledge and \textit{Zero-Shot Degradation}, \ie, the loss of accuracy on unseen tasks.
A common way to mitigate these effects is through Knowledge Distillation (KD), which serves as a regularizer during optimization.

Unlike the main competitor~\cite{yu2024select}, who requires two teacher models to fully regulate the learning process (the model learned during the previous task $g^{k-1}$ and the reference model at the beginning of continual optimization $g^{0}$), 
our geometry-aware (Sec. \ref{subsec:geodesic}) approach only requires $g^{k-1}$ to effectively tackle both \textit{catastrophic forgetting} and \textit{zero-shot degradation}.

Moreover, to enhance network plasticity and reduce the degradation effect brought by the Cross-Entropy loss, the current task is learned within a projected low-rank subspace (Sec. \ref{subsec:subspace}). Figure~\ref{fig:arch} summarizes our pipeline.

\subsection{Subspace Learning for Plasticity Restoration}\label{subsec:subspace}
Knowledge distillation strategies have the drawback of reducing model plasticity and, consequently, performance on novel tasks. A key reason for this is that the regularization is applied uniformly across the entire high-dimensional manifold.

We introduce the idea that an effective way to handle this issue is to constrain optimization for the new task to a specific subspace, leaving most of the original manifold intact.
This allows the new task to be learned flexibly within that subspace, while the regularization remains largely unchanged elsewhere, thus reducing potential conflicts between the task objective and the regularization. This insight leads to the first main contribution of our work: we propose using subspace learning to restrict parameter updates to a low-rank subspace of the vision language model. 
More specifically, at task $k$, we exploit a learnable low-rank projector $U_k \in \mathbb{R}^{d\times r}, \; r \ll d$ (in our experiments we used $r=144$) to map the VLM embeddings into \textit{task-specific} ($\mathbf{f}_\parallel = U_k U_k^\top \mathbf{f}$) and \textit{task-irrelevant} ($\mathbf{f}_\perp = \mathbf{f} - \mathbf{f}_\parallel$) subspaces, allowing us to supervise the two regions of the manifold in complementary ways. 
During training, we force $U_k$ to be a basis for the subspace, \ie, $U_k^\top U_k = I \in \mathbb{R}^{r\times r}$\footnote{$U_k$ is computed via reparameterization. We optimize a randomly initialized proxy matrix $A \in \mathbb{R}^{d\times r}$, decompose it via QR-decomposition in reduced mode, and use the orthonormal $Q$ as $U_k$. The linearity of the decomposition guarantees gradient flow.}.
Note that the low-rank projection is used exclusively during training, as it is only required for computing the loss. During inference, this projection is removed, and the full-dimensional embeddings are employed instead, allowing for a seamless transition between the MTIL and MCIL configurations.
Formally, we first compute the vector of class probabilities  in the low-rank subspace for a given sample $x$ as:
\begin{equation}
\mathbf{s}_\parallel = \text{softmax}([\left<U_k U_k^\top \mathbf{f}_i, U_k U_k^\top g_t(c)\right> \; \forall c \in \mathcal{C}])\;,
\end{equation}
where we re-normalize the projected embeddings before use for consistency with the previous formulation.
The vector is then used for the subspace learning objective, which is defined as the cross entropy between the vectors $\mathbf{s}_\parallel$ and the one-hot encoded labels:
\begin{equation}
    \mathcal{L}_{\text{sub}} = \frac{1}{N}\sum_{j=1}^N \text{CE}(\mathbf{s}_{j,\parallel}, \mathbf{y}_j)\;.
\end{equation}

\subsection{Geodesic Distance for Strong Stability}\label{subsec:geodesic}
We previously mentioned that using a geometric-aware optimization objective helps reduce catastrophic forgetting and zero-shot degradation. This stems from the topology that the CLIP latent space has attained during training (high-dimensional sphere). This insight allows us to recognize the geodesic distance as a suitable metric, as it respects the inherent structure of CLIP's latent space. 

More specifically, while the $L2$ distance (proportional to the cosine distance in normalized embeddings) is the standard choice to compare CLIP features, it does not perfectly match the geometry of CLIP's image embeddings, since they lie on the surface of a high-dimensional sphere due to vector normalization~\cite{mei2025geommgeodesicperspectivemultimodal,kang2025clip}.

In contrast, the geodesic distance serves as a more appropriate measure, aligning with the pretrained latent space geometry of the CLIP model and potentially preserving its zero-shot abilities in continual learning scenarios.

Given an input sample $\mathbf{x}$ and two  encoders $g_i^k$ and $g_i^{k-1}$, the geodesic loss can be expressed as:
\begin{equation}
    \mathcal{L}_{\text{geo}} = \cos^{-1}\left(\left<g_i^{k}(\mathbf{x}), g_i^{k-1}(\mathbf{x})\right>\right)\;.
    \label{eq:kd}
\end{equation}
Intuitively, rather than measuring distance along straight lines through the high-dimensional sphere, the metric computes distance along the shortest path on the curved surface, preserving the underlying geometry imparted by the cosine distance during training. For this reason, we introduce the concept of using this metric to compare descriptors.

\subsection{Learning Objective}
Furthermore, to tackle the stability-vs-plasticity trade-off, we also leverage the learned projector $U_k$ at each task $\mathcal{T}^k$ to decompose the image embeddings into two orthogonal subspaces and apply knowledge distillation separately to both.

More specifically, given the embedding $\mathbf{f}^k$ from $g^k$ and $\mathbf{f}^{k-1}$ from $g^{k-1}$, we compute the knowledge distillation loss as the sum of the two geodesic distances in the task-specific and task-irrelevant subspaces:
\begin{equation}
    \mathcal{L}_{KD} = \frac{1}{N}\sum_{j=1}^N \left[ \mathcal{L}_{\text{geo}}(\mathbf{f}_{\parallel,j}^k, \mathbf{f}_{\parallel,j}^{k-1}) + \mathcal{L}_{\text{geo}}(\mathbf{f}_{\perp,j}^k, \mathbf{f}_{\perp,j}^{k-1}) \right]\;.
\end{equation}
Note that computing the metric on the separate projected subspaces leads to different results than computing it on the original vectors (see the Suppl. Mat. Section 1.3).

These considerations lead us to the total objective function, which is a convex combination of the global Cross-Entropy loss $\mathcal{L}_{CE}$, the subspace learning loss $\mathcal{L}_{\text{sub}}$, and the decomposed distillation loss $\mathcal{L}_{KD}$:
\begin{equation}
    \mathcal{L} = \mathcal{L}_{CE} + \alpha\mathcal{L}_{\text{sub}} + \beta\mathcal{L}_{KD}\;,
\end{equation}
where $\alpha=0.5$ and $\beta=3$ are hyperparameters that control the trade-off between plasticity and stability. Please refer to Sec. \ref{sec:ablation_main} for details. 

%%%%%%%%%%%%%%%%%%%%%%%%%%%%%%%%%%%%%%%%%%%%%%%%%%%%%%%%%%%%%%%%%%%%%%%%%%%%%%%%%%%%%%%%%%%%%%%%%%%%%%
\section{Experimental Evaluation}
\label{sec:experiments}
This section presents the experimental evaluation of our approach. We start by detailing the setup and the datasets used for the experiments. Then, we present the experimental comparison with the state-of-the-art and, finally, some ablation data analyzing the impact of the different components and parameters.

\subsection{Experimental Setup}
For our experimental evaluation, we use the CLIP architecture~\cite{radford2021learning} implemented by \texttt{open\_clip}~\cite{ilharco2021openclip} as the reference VLM. 
The image encoder $g_i$ is a ViT-B/16 \cite{dosovitskiy2020image}, while the text encoder $g_t$ is a $12$-layer decoder-only transformer network with a latent space dimension of $512$ and a context length of $77$ tokens.

In our experiments, only the vision encoder is optimized, while the text encoder remains frozen. The encoder $g_i$ is optimized using AdamW, with a peak learning rate of $1\times 10^{-5}$ decayed to $0$ via a cosine annealing scheduler, and the weight decay is set to $5\times 10^{-4}$. 
In each incremental step, training lasts for $1000$ iterations or $10$ epochs, whichever comes first. A full training on all 8 datasets lasts around 96 minutes on an NVIDIA 4090. 

Following the procedure of \cite{zheng2023preventing, yu2024select}, we select the same $100K$ unlabeled images from ImageNet~\cite{deng2009imagenet} to be used as the reference dataset for knowledge distillation.

\subsection{Datasets}
Following previous work, we evaluate our method in a continual setting that includes a sequence of 8 fine-grained classification datasets in both MTIL and MCIL benchmarks.
More specifically, we use FGVC-Aircraft~\cite{maji2013fine} (containing $100$ airplane models), DTD~\cite{cimpoi2014describing} (containing $47$ patterns and textures), EuroSAT~\cite{helber2019eurosat} (containing $10$ types of satellite imagery), Flowers-102~\cite{nilsback2008automated} (containing $102$ species of flowers), Food-101~\cite{bossard2014food} (containing $101$ classes of foodstuffs), Oxford-Pets~\cite{parkhi2012cats} (containing $37$ species of cats and dogs), Stanford-Cars~\cite{krause20133d} (containing $196$ car models), and UCF-101~\cite{khurram2012dataset} (containing $101$ action recognition classes). 

Since task ordering can substantially influence continual learning performance \cite{bell2022effect, li2025optimal}, we did not restrict our analysis to only two sequences, as in \cite{zheng2023preventing}. Instead, we adopted the validation protocol proposed by \cite{yu2024select}, constructing 8 sequences of tasks $\mathcal{S}^k, k=1,..,8$ as follows:
\begin{equation}
    \mathcal{S}^k = (\mathcal{T}^{k (\!\!\!\!\!\!\mod K)}, \mathcal{T}^{k+1 (\!\!\!\!\!\!\mod K)}, \cdots, \mathcal{T}^{k+K-1 (\!\!\!\!\!\!\mod K)})\;,
\end{equation}
where $K=8$ is the total number of tasks. 
Essentially, we set $\mathcal{S}^1$ as the original ordering (\ie, FGVC-Aircraft, DTD, EuroSAT, Flowers-102, Food-101, Oxford-Pets, Stanford-Cars, UCF-101) and construct the other 7 by rotating the sequence to the left one task at a time. 
Details of all the 8 sequences are provided in the supplementary material.

\begin{figure}[t]
    \centering
    \includegraphics[width=\linewidth]{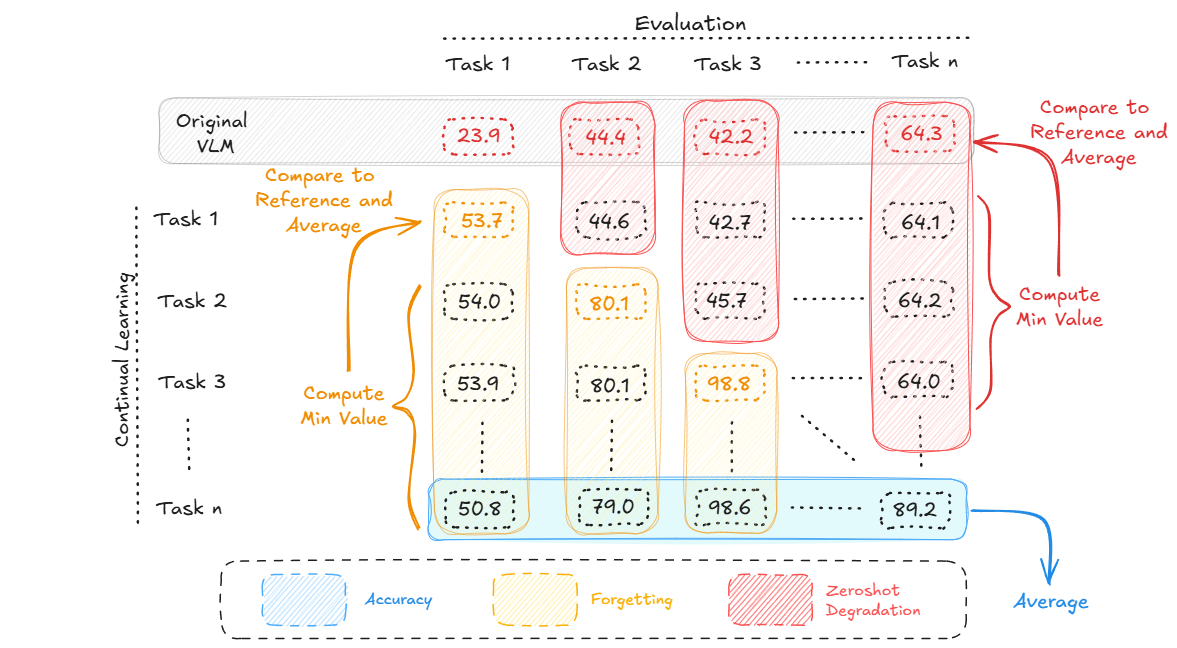}
    \caption{Schematic overview of how the metrics are computed. \textit{Accuracy} denotes the mean performance over all tasks once Continual Learning has finished. \textit{Catastrophic Forgetting} quantifies the average, worst-case loss in performance on previously learned tasks, relative to supervised training on each task. \textit{Zero-Shot Degradation} captures the average, worst-case drop in performance on unseen future tasks, relative to the accuracy of the pretrained model before the start of continual learning.}
    \label{fig:metrics}
\end{figure}

\subsection{Metrics}
We consider three metrics in this paper: accuracy, catastrophic forgetting, and zero-shot degradation. Following \cite{yu2024select, chaudhry2018riemannian, chaudhry2019tiny, lopez2017gradient}, we adopt the evaluation procedures illustrated in Fig.~\ref{fig:metrics}: accuracy is computed as the average of per-task accuracies after training on the final task; catastrophic forgetting measures the average maximum performance drop across all previous tasks; and zero-shot degradation evaluates the average maximum performance drop on unseen tasks.

%%%%%%%%%%%%%%%%%%%%%%%%%%%%%%%%%%%%%%%%%%%%%%%%%%%%%% MTIL benchmark
\begin{table*}[t]
\centering
\small
\setlength{\tabcolsep}{1.8pt}
\renewcommand{\arraystretch}{1.1}
\caption{Task-Specific evaluation (MTIL task). During inference, each image is compared only against the class embeddings of its task. We report percent Accuracy ($\uparrow$), Catastrophic Forgetting ($\downarrow$), and Zero-Shot Degradation (Z. S. Deg., $\downarrow$). \textbf{Best} in bold.}
\label{tab:main_results_MTIL}
\begin{tabular}{clccccccccc}
\toprule
\toprule
& Method & $\mathbf{S^1}$ & $\mathbf{S^2}$ & $\mathbf{S^3}$ & $\mathbf{S^4}$ & $\mathbf{S^5}$ & $\mathbf{S^6}$ & $\mathbf{S^7}$ & $\mathbf{S^8}$ & \textbf{Mean} \\
\midrule

%\multicolumn{10}{l}{\textbf{Accuracy} ($\uparrow$)} \\
\multirow{7}{*}{\rotatebox{90}{\textbf{Accuracy} ($\rightarrow$)}} & Continual FT      & 76.16 & 76.24 & 78.03 & 68.69 & 76.64 & 75.44 & 72.71 & 77.45 & 75.17 \\
& LwF \cite{li2017learning}   & 76.78 & 80.45 & 80.65 & 77.52 & 79.64 & 79.45 & 77.31 & 78.70 & 78.81 \\
& iCaRL  \cite{rebuffi2017icarl}  & 77.99 & 79.77 & 79.93 & 76.66 & 79.26 & 79.08 & 77.06 & 78.61 & 78.55 \\
& ZSCL\cite{zheng2023preventing}  & 81.89 & 83.98 & 84.30 & 83.49 & 83.41 & 82.38 & 81.92 & 81.97 & 82.92 \\
& MoE-Adapters \cite{yu2024boosting}  & 82.71 & 80.74 & 81.15 & 83.97 & 83.68 & 83.68 & 82.73 & 79.68 & 82.29 \\

& GIFT\cite{wu2025synthetic} & 81.93 & 84.01 & 84.10 & 83.67 & 84.11 & 84.25 & 82.54 & 83.10 & 83.46\\
& SnD\cite{yu2024select}& 84.48 & 84.92 & 84.97 & 84.89 & 85.50 & 85.07 & 85.02 & 84.52 & 84.92 \\

\cline{2-11}
% & \textbf{Ours}     & \textbf{85.72} &  \textbf{86.03} & \textbf{86.04} & \textbf{86.06} & \textbf{86.34} & \textbf{85.77} & \textbf{85.78} & \textbf{85.73} & \textbf{85.93} \\
% & \textbf{Ours-new} & 85.75 & 86.07 & 85.91 & 85.96 & 86.06 & 85.87 & 85.71 & 85.77 & 85.89\\
& \textbf{Ours} & \textbf{85.74} & \textbf{86.11} & \textbf{86.06} & \textbf{85.99} & \textbf{86.05} & \textbf{85.78} & \textbf{85.70} & \textbf{85.68} & \textbf{85.89} \\
\midrule
\midrule

% \multicolumn{10}{l}{\textbf{Forgetting} ($\downarrow$)} \\
\multirow{7}{*}{\rotatebox{90}{\textbf{Forgetting} ($\leftarrow$)}} & Continual FT     & 10.98 & 10.60 & 8.80 & 19.17 & 10.11 & 11.95 & 15.19 & 9.48 & 12.04 \\
& LwF \cite{li2017learning}  & 10.38 & 6.52 & 6.37 & 10.22 & 7.99 & 7.70 & 10.41 & 8.91 & 8.56 \\
& iCaRL \cite{rebuffi2017icarl} & 8.42 & 7.00 & 6.45 & 10.21 & 7.03 & 7.33 & 9.68 & 8.23 & 8.04 \\
& ZSCL \cite{zheng2023preventing} & 4.67 & 2.35 & 2.13 & 2.97 & 3.15 & 4.28 & 4.89 & 4.70 & 3.64 \\
& MoE-Adapters\cite{yu2024boosting} & 2.74 & 4.71 & 4.28 & 1.15 & 1.50 & 1.60 & 2.94 & 2.77 & 2.71 \\

& GIFT\cite{wu2025synthetic} & 5.81 & 3.08 & 3.35 & 3.81 & 3.17 & 3.26 & 4.82 & 4.31 & 3.95\\
& SnD \cite{yu2024select} & 1.70 & 1.16 & 0.89 & 1.04 & 0.59 & 1.34 & 1.12 & 1.79 & 1.20 \\
\cline{2-11}

% & \textbf{Ours-last}         & \textbf{0.79} &  \textbf{0.42} & \textbf{0.51} & \textbf{0.71} & \textbf{0.48} & \textbf{0.90} & \textbf{0.90} & \textbf{0.77} &  \textbf{0.69}\\
% & \textbf{Ours-max}         & 0.75 & 0.43 & 0.47 & 0.52 & 0.56 & 0.85 & 1.06 & 0.74 & 0.67\\
& \textbf{Ours}         & \textbf{0.81} & \textbf{0.40} & \textbf{0.44} & \textbf{0.65} & \textbf{0.48} & \textbf{0.83} & \textbf{0.96} &  \textbf{0.82} & \textbf{0.67} \\
\midrule
\midrule

% \multicolumn{10}{l}{\textbf{Zero-Shot degradation} ($\downarrow$)} \\
\multirow{7}{*}{\rotatebox{90}{\textbf{Z. S. Deg.} ($\leftarrow$)}} & Continual FT      & 24.81 & 23.58 & 19.54 & 16.46 & 22.22 & 19.02 & 19.54 & 24.02 & 21.15 \\
& LwF \cite{li2017learning} & 10.75 & 10.23 & 8.63 & 8.25 & 12.02 & 10.33 & 8.98 & 11.01 & 10.03 \\
& iCaRL\cite{rebuffi2017icarl}  & 13.77 & 12.68 & 11.28 & 12.14 & 13.20 & 13.20 & 13.09 & 14.01 & 12.92 \\
& ZSCL \cite{zheng2023preventing} & 3.44 & 3.94 & 4.02 & 2.85 & 3.79 & 2.31 & 1.86 & 1.84 & 3.00 \\
& MoE-Adapters\cite{yu2024boosting} & 1.62 & 2.58 & 1.04 & 2.37 & 4.31 & 3.05 & 1.77 & 0.63 & 2.17 \\
& GIFT\cite{wu2025synthetic} & 0.80 & 0.82 & 1.58 & 3.86 & 4.23 & 3.88 & 3.56 & 4.39 & 2.89\\
& SnD\cite{yu2024select} & 1.55 & 2.04 & 1.21 & 1.92 & 2.79 & 2.18 & 1.90 & 2.08 & 1.96 \\

\cline{2-11}
% & \textbf{Ours-last}         & \textbf{-0.14} & \textbf{0.29} & \textbf{0.53} & \textbf{-0.10} & \textbf{-0.39} & \textbf{-0.41} & \textbf{-0.37} & \textbf{-0.45} & \textbf{-0.13} \\
% & \textbf{Ours-max}         & 0.35 & 0.33 & 0.70 & 0.54 & 0.59 & 0.60 & 0.62 & 0.15 & 0.49 \\
& \textbf{Ours}         & \textbf{0.36} & \textbf{0.19} & \textbf{0.68} & \textbf{0.58} & \textbf{0.67} & \textbf{0.64} & \textbf{0.55} & \textbf{0.12} & \textbf{0.47}\\
\bottomrule
\bottomrule
\end{tabular}
\end{table*}
%%%%%%%%%%%%%%%%%%%%%%%%%%%%%%%%%%%%%%%%%%%%%%%%%%%%%%

\subsection{Baseline Methods}
We compare our approach with several recent methods focusing on CLIP. As a baseline, we include basic Continual Fine-Tuning (FT), which naively fine-tunes CLIP on each task sequentially without regularization. This method serves as a lower-bound reference for all comparisons. We also compare it with classic continual learning methods (LwF\cite{li2017learning} and iCaRL\cite{rebuffi2017icarl}) and with recent state-of-the-art techniques such as  ZSCL~\cite{zheng2023preventing}, MoE-Adapters~\cite{yu2024boosting}, SnD~\cite{yu2024select} and GIFT~\cite{wu2025synthetic}.

\begin{table*}[t]
\centering
\small
\renewcommand{\arraystretch}{1.1}
\setlength{\tabcolsep}{2.5pt}
\caption{Task-Agnostic evaluation (MCIL task). During inference, each image is compared against all 694 textual class embeddings. We report percent Accuracy ($\uparrow$), Catastrophic Forgetting ($\downarrow$), and Zero-Shot Degradation (Z. S. Deg., $\downarrow$). \textbf{Best} in bold.}
\label{tab:main_results_MCIL}
\begin{tabular}{clccccccccc}
\toprule
\toprule
& Method & $\mathbf{S^1}$ & $\mathbf{S^2}$ & $\mathbf{S^3}$ & $\mathbf{S^4}$ & $\mathbf{S^5}$ & $\mathbf{S^6}$ & $\mathbf{S^7}$ & $\mathbf{S^8}$ & \textbf{Mean} \\
\midrule
% \multicolumn{10}{l}{\textbf{Accuracy} ($\uparrow$)} \\
\multirow{6}{*}{\rotatebox{90}{\textbf{Accuracy} ($\rightarrow$)}} & Continual FT      & 75.17 & 75.13 & 76.01 & 67.17 & 75.54 & 74.47 & 71.66 & 76.40 & 73.94\\
& LwF \cite{li2017learning}   & 74.14 & 77.44 & 77.94 & 74.89 & 77.30 & 77.43 & 75.50 & 76.51 & 76.39\\
& iCaRL  \cite{rebuffi2017icarl}  & 76.97 & 78.82 & 78.57 & 75.43 & 78.08 & 78.10 & 75.70 & 77.52 & 77.40\\
& ZSCL\cite{zheng2023preventing}  & 80.49 & 82.54 & 82.99 & 82.08 & 82.17 & 80.99 & 80.30 & 80.09 & 81.46 \\
& GIFT\cite{wu2025synthetic} & 79.99 & 81.68 & 81.95 & 83.08 & 83.48 & 83.27 & 81.58 & 81.27 & 82.04 \\
& SnD\cite{yu2024select} & 83.35 & 83.57 & 83.88 & 83.70 & 84.46 & 83.82 & 83.89 & 83.43 & 83.76\\

\cline{2-11}
%& \textbf{Ours}     & \textbf{85.07} & \textbf{85.18} & \textbf{85.36} & \textbf{85.27} & \textbf{85.31} & \textbf{84.99} & \textbf{85.05} & \textbf{84.92}  & \textbf{85.14} \\
& \textbf{Ours} & \textbf{85.00} & \textbf{85.27} & \textbf{85.37} & \textbf{85.27} & \textbf{85.28} & \textbf{85.05} & \textbf{84.97} & \textbf{84.96} & \textbf{85.15} \\
\midrule
\midrule

% \multicolumn{10}{l}{\textbf{Forgetting} ($\downarrow$)} \\
\multirow{6}{*}{\rotatebox{90}{\textbf{Forgetting} ($\leftarrow$)}} & Continual FT     & 11.17 & 10.89 & 10.16 & 20.12 & 10.57 & 12.14 & 15.62 & 9.80 & 12.56 \\
& LwF \cite{li2017learning}  & 9.56 & 6.38 & 6.93 & 11.09 & 8.37 & 7.69 & 10.24 & 8.44 & 8.59\\
& iCaRL \cite{rebuffi2017icarl} & 8.43 & 6.90 & 6.83 & 10.69 & 7.09 & 7.37 & 10.17 & 8.66 & 8.27\\
& ZSCL \cite{zheng2023preventing} & 4.21 & 1.41 & 2.08 & 3.32 & 2.85 & 4.39 & 5.22 & 5.13 & 3.58 \\
& GIFT\cite{wu2025synthetic} & 6.85 & 4.52 & 5.07 & 3.51 & 3.42 & 3.80 & 5.19 & 5.46 & 4.73 \\
& SnD \cite{yu2024select} & 1.92 & 1.53 & 0.97 & 1.14 & \textbf{0.58} & 1.55 & 1.29 & 1.81 & 1.35\\

\cline{2-11}
%& \textbf{Ours-last}         & \textbf{0.93} & \textbf{0.71} & \textbf{0.68} & \textbf{0.80} & \textbf{0.44} & \textbf{1.05} & \textbf{1.00} & \textbf{1.05} & \textbf{0.83} \\
& \textbf{Ours} & \textbf{1.05} & \textbf{0.63} & \textbf{0.58} & \textbf{0.77} & 0.68 & \textbf{0.99} & \textbf{1.14} & \textbf{1.02} & \textbf{0.86} \\
\midrule
\midrule
%\multicolumn{10}{l}{\textbf{Zero-Shot degradation} ($\downarrow$)} \\
\multirow{6}{*}{\rotatebox{90}{\textbf{Z. S. Deg.} ($\leftarrow$)}} & Continual FT      & 24.54 & 24.10 & 19.53 & 17.60 & 21.96 & 18.92 & 20.26 & 24.31 & 21.40\\
& LwF \cite{li2017learning} & 11.94 & 11.82 & 8.27 & 9.99 & 13.36 & 11.47 & 10.95 & 12.63 & 11.30\\
& iCaRL\cite{rebuffi2017icarl}  & 13.02 & 12.78 & 10.89 & 11.81 & 12.74 & 12.87 & 11.92 & 13.34 & 12.42\\
& ZSCL \cite{zheng2023preventing} & 3.59 & 4.71 & 4.17 & 2.81 & 3.55 & 1.97 & 1.47 & 2.30 & 3.07\\
& GIFT\cite{wu2025synthetic} & 2.71 & 1.87 & 1.12 & 2.66 & 3.56 & 3.78 & 3.15 & 3.97 & 2.85 \\
& SnD\cite{yu2024select} & 1.44 & 1.80 & 1.01 & 1.53 & 2.17 & 1.80 & 1.65 & 1.82 & 1.65\\

\cline{2-11}
% & \textbf{Ours-last}         & \textbf{0.26} & \textbf{0.55} & \textbf{0.52} & \textbf{0.30} & \textbf{0.21} & \textbf{0.11} & \textbf{0.10} & \textbf{-0.04}& \textbf{0.25} \\
& \textbf{Ours} & \textbf{0.47} & \textbf{0.38} & \textbf{0.67} & \textbf{0.69} & \textbf{0.85} & \textbf{0.85} & \textbf{0.76} & \textbf{0.34} & \textbf{0.63} \\
\bottomrule
\bottomrule
\end{tabular}
\end{table*}

\subsection{Main Results}
In Tables \ref{tab:main_results_MTIL} and \ref{tab:main_results_MCIL}, we report the quantitative results attained by TASSO in the Multi Domain Task-Incremental Learning (MTIL) and Class-Incremental Learning (MCIL) settings, respectively. More specifically, the two tables report Accuracy (higher-is-better), Catastrophic Forgetting (lower-is-better), and Zero-Shot Degradation (lower-is-better) in the MTIL and MCIL settings.

Starting from the easier MTIL setting (where the task and the corresponding label space are known), the results in Table~\ref{tab:main_results_MTIL} show that our method achieves noticeable gains over competing approaches. 

More specifically, TASSO achieves the best accuracy across all orderings, surpassing the best competitor (\ie, \cite{yu2024select}) by an average of $\approx\!1\%$, achieving an impressive absolute accuracy of $85.89\%$. Even more remarkable is the reduction in Catastrophic Forgetting, which is just $0.67\%$, compared to $1.2\%$ of the best competitor and much larger values for all the others. Notice how forgetting preservation is stable across all tasks, never surpassing the $1\%$ threshold (the worst result is a $0.96\%$ in sequence $\mathbf{S^7}$).

Nevertheless, our approach shines in Zero-Shot Degradation, achieving an average of just $0.47\%$, 4 times lower than the closest competitors \cite{yu2024select,yu2024boosting}, who hover around $2\%$. This indicates that TASSO is generally able to retain the existing knowledge within the VLM, preserving accuracy for future unseen tasks.

The behavior is also consistent across all task orderings (\ie, $\mathbf{S^k}$ in the Tables), with values ranging from an impressive $0.19\%$ to a maximum of just $0.68\%$.
These results also support the hypothesis that some datasets provide information useful for distinguishing samples from other tasks. 

For example, the DTD dataset (which appears first in $S^2$, leading to the best performance, and last in $S^3$, leading to the worst) contains several different patterns and textures. Tuning the CLIP architecture on it may help it better understand and distinguish images from a general point of view. This capability is then reflected in higher accuracy on later tasks (\ie, those measured by the Zero-Shot Degradation metric).

Table~\ref{tab:main_results_MCIL} shows results from the more challenging MCIL benchmark, where the selected task is unknown, and the class must be recognized among all those learned in the incremental steps (totaling $694$ classes). 

As expected, the results are slightly lower in absolute terms, reaching an average accuracy of $85.15\%$. However, the improvement compared to the competitors actually increases to $\approx\!1.4\%$ and is consistent across all sequences (TASSO has the best accuracy on all 8 considered sequences).

As before, Catastrophic Forgetting is significantly reduced, reaching an average of $0.86\%$, with only $\mathbf{S^1}$, $\mathbf{S^7}$, and $\mathbf{S^8}$ surpassing the $1\%$ threshold. TASSO outperforms competitors also in this setting with a gain of around $0.5\%$ on the best competitors being the best in 7 out of 8 cases.

The most remarkable improvements, however, are still found in Zero-Shot Degradation, where our approach averages $0.63\%$, corresponding to a reduction of $1\%$ compared to the closest alternative. The degradation is almost 3 times smaller than \cite{yu2024select}, which is the best competitor, while the improvement is also consistent across all sequences, all of which have a degradation below $1\%$.

%%%%%%%%%%%%%%%%%%%%%%%%%%%%%%%%% Cosine similarity figures
\begin{figure*}[t]
    \centering
    \begin{subfigure}[b]{0.22\textwidth}
        \includegraphics[width=\linewidth]{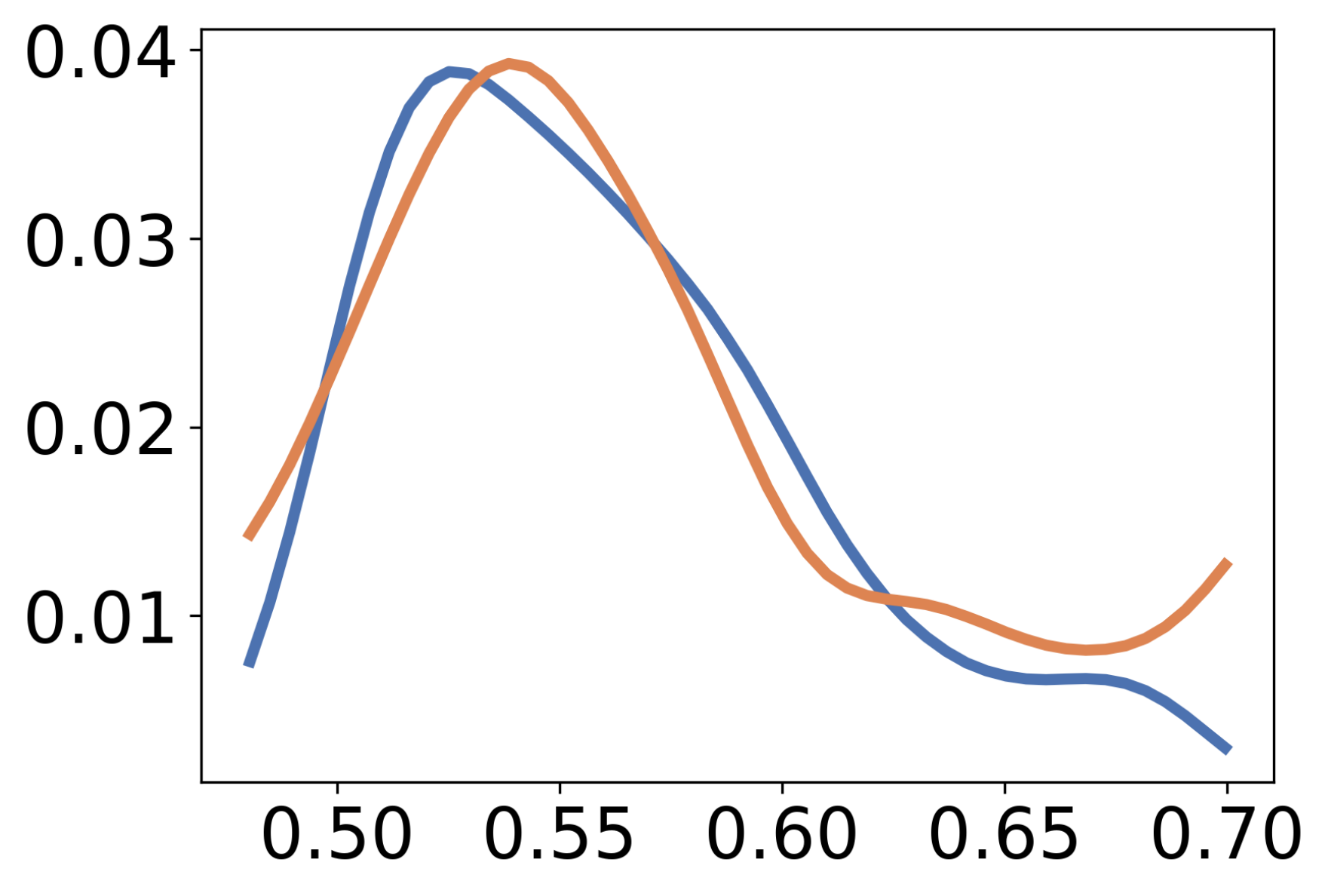}
        \caption{FGVCAircraft}
        \label{fig:fgvc-aircraft}
    \end{subfigure}
    \hfill
    \begin{subfigure}[b]{0.22\textwidth}
        \includegraphics[width=\linewidth]{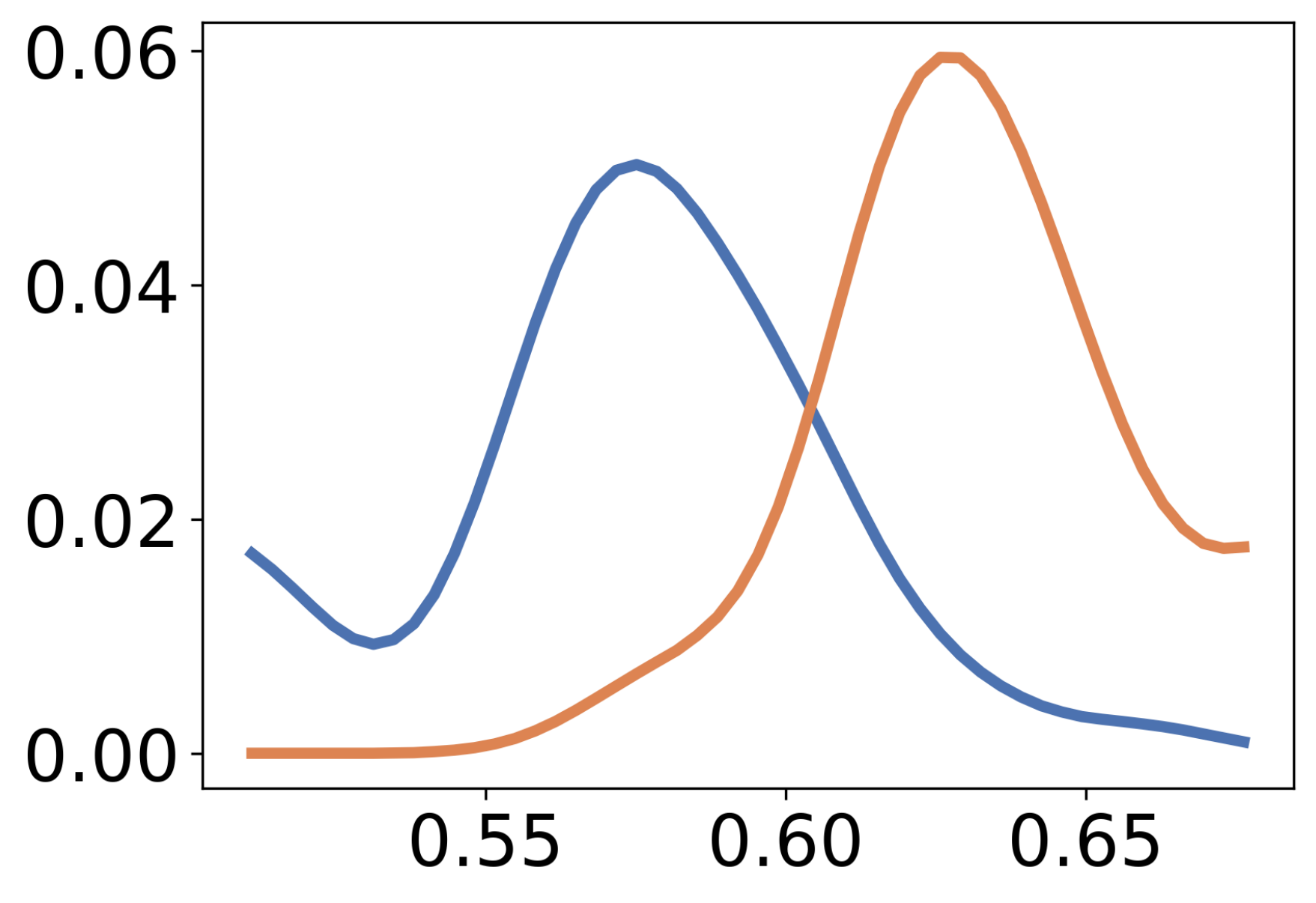}
        \caption{DTD}
        \label{fig:dtd}
    \end{subfigure}
    \hfill
    \begin{subfigure}[b]{0.22\textwidth}
        \includegraphics[width=\linewidth]{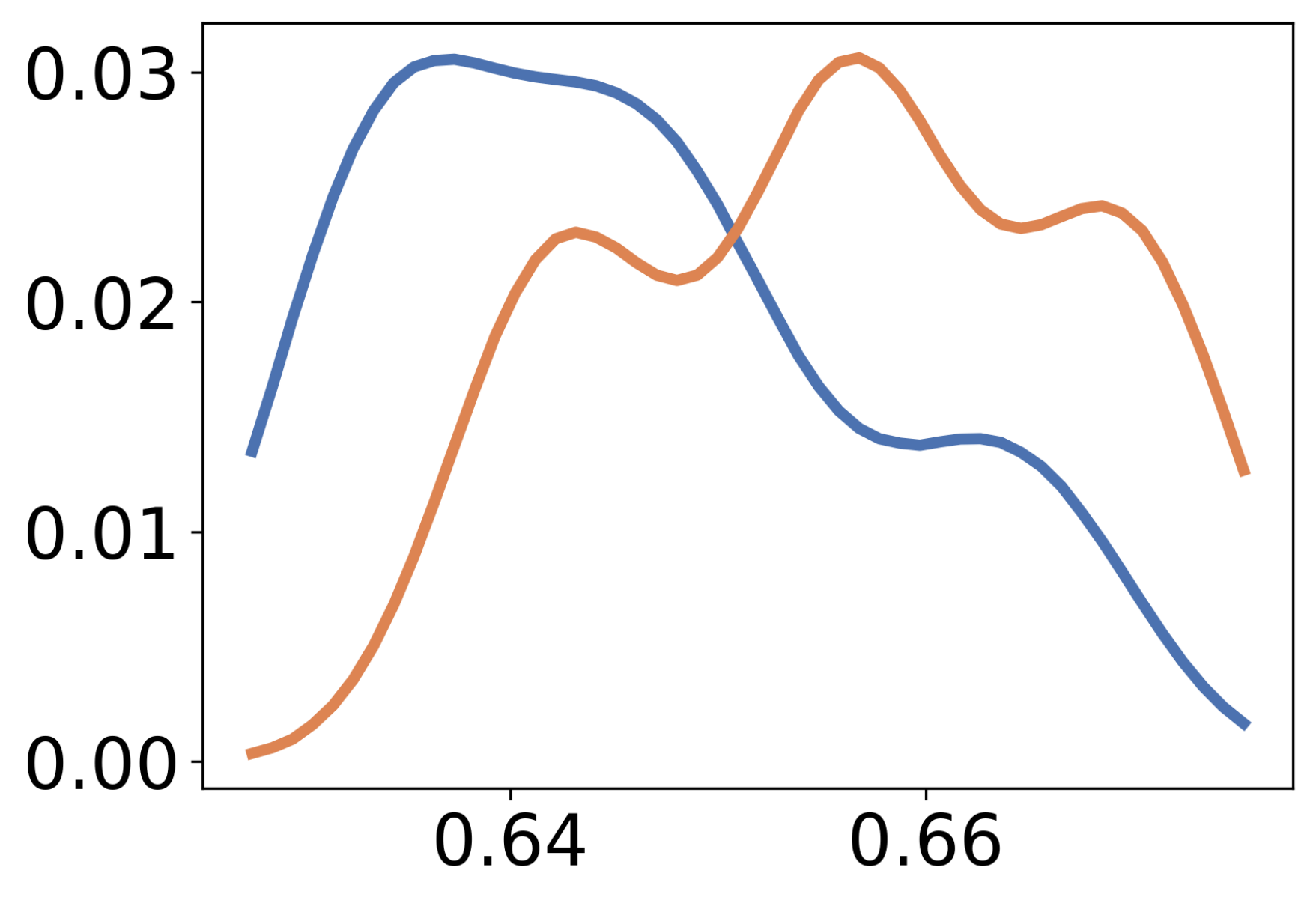}
        \caption{EuroSat}
        \label{fig:eurosat}
    \end{subfigure}
    \hfill
    \begin{subfigure}[b]{0.22\textwidth}
        \includegraphics[width=\linewidth]{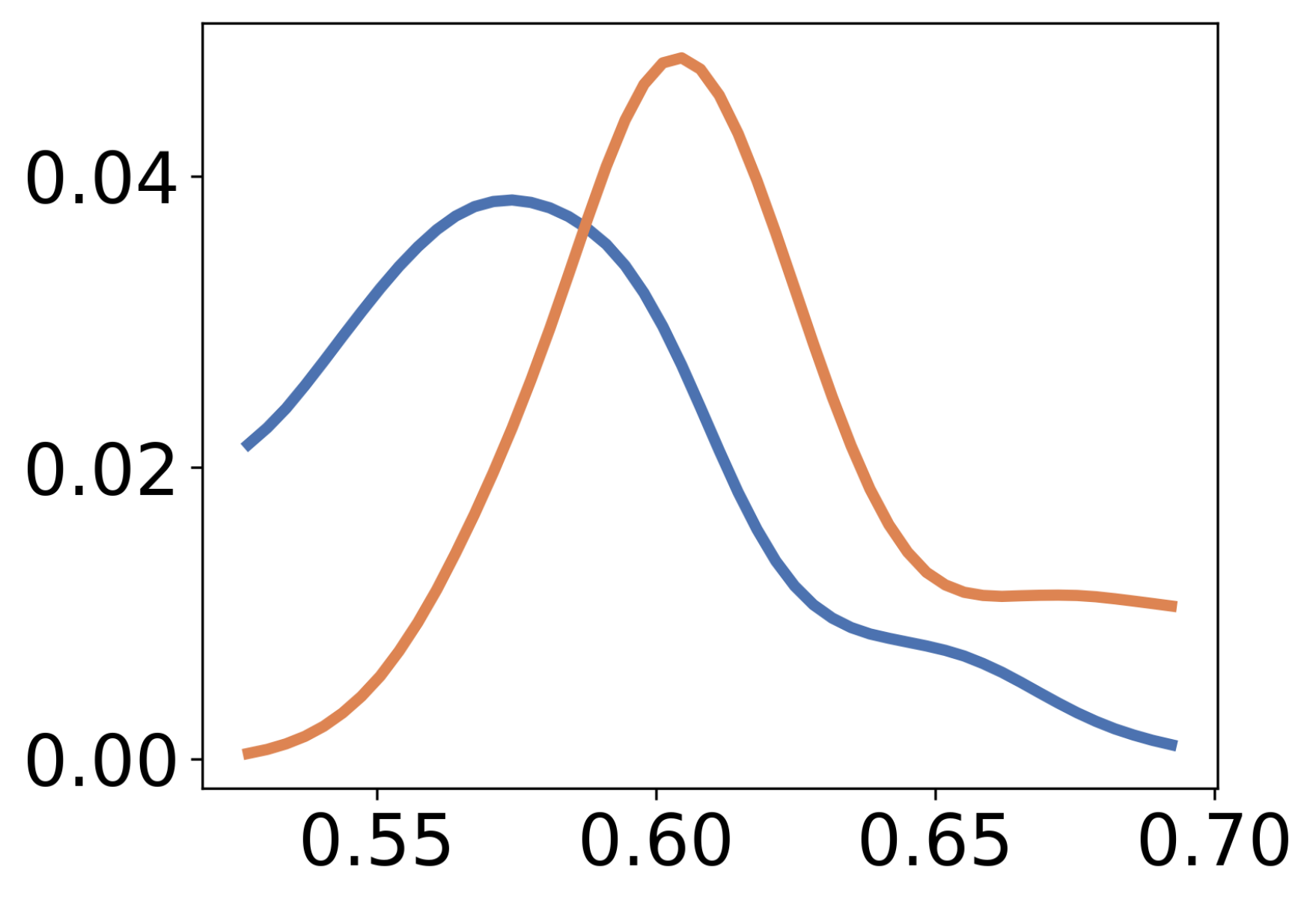}
        \caption{Flowers-102}
        \label{fig:flowers-102}
    \end{subfigure}
    %%%%%%%%%%%%%%%%%%%%%%%%%%%%%%%%%% second row
    \begin{subfigure}[b]{0.22\textwidth}
        \includegraphics[width=\linewidth]{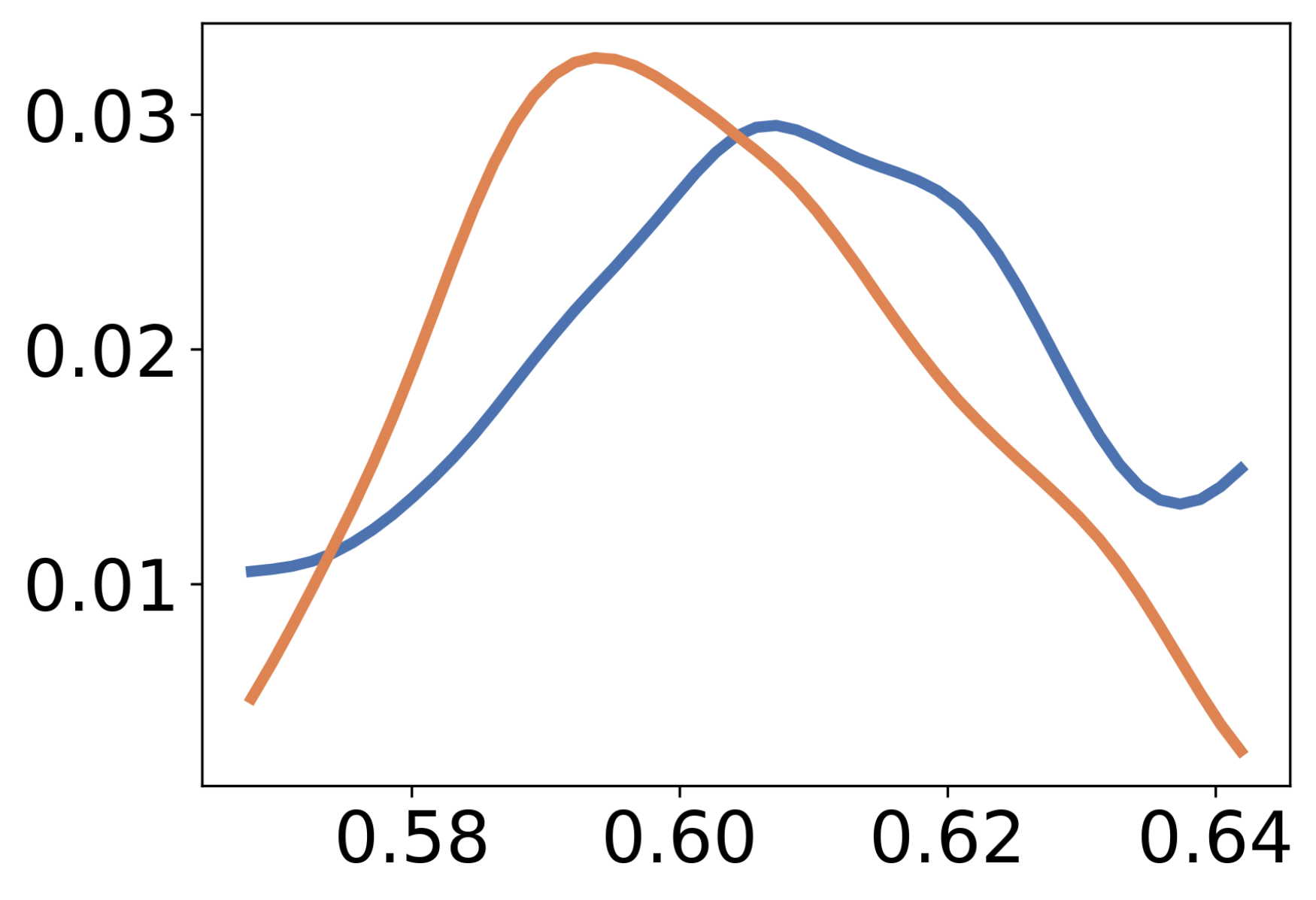}
        \caption{Food101}
        \label{fig:food-101}
    \end{subfigure}
    \hfill
    \begin{subfigure}[b]{0.22\textwidth}
        \includegraphics[width=\linewidth]{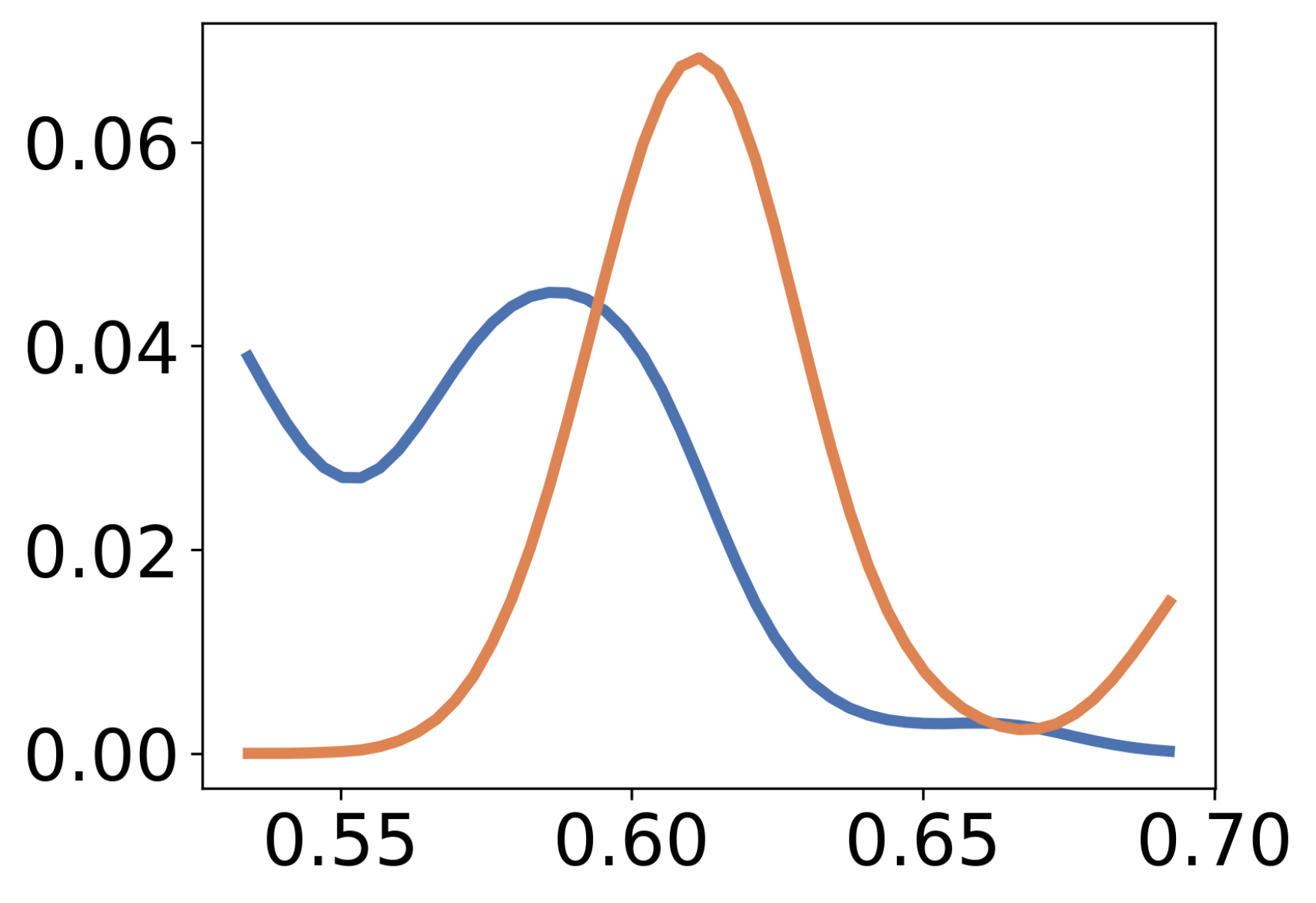}
        \caption{OxfordPets}
        \label{fig:oxford-pets}
    \end{subfigure}
    \hfill
    \begin{subfigure}[b]{0.22\textwidth}
        \includegraphics[width=\linewidth]{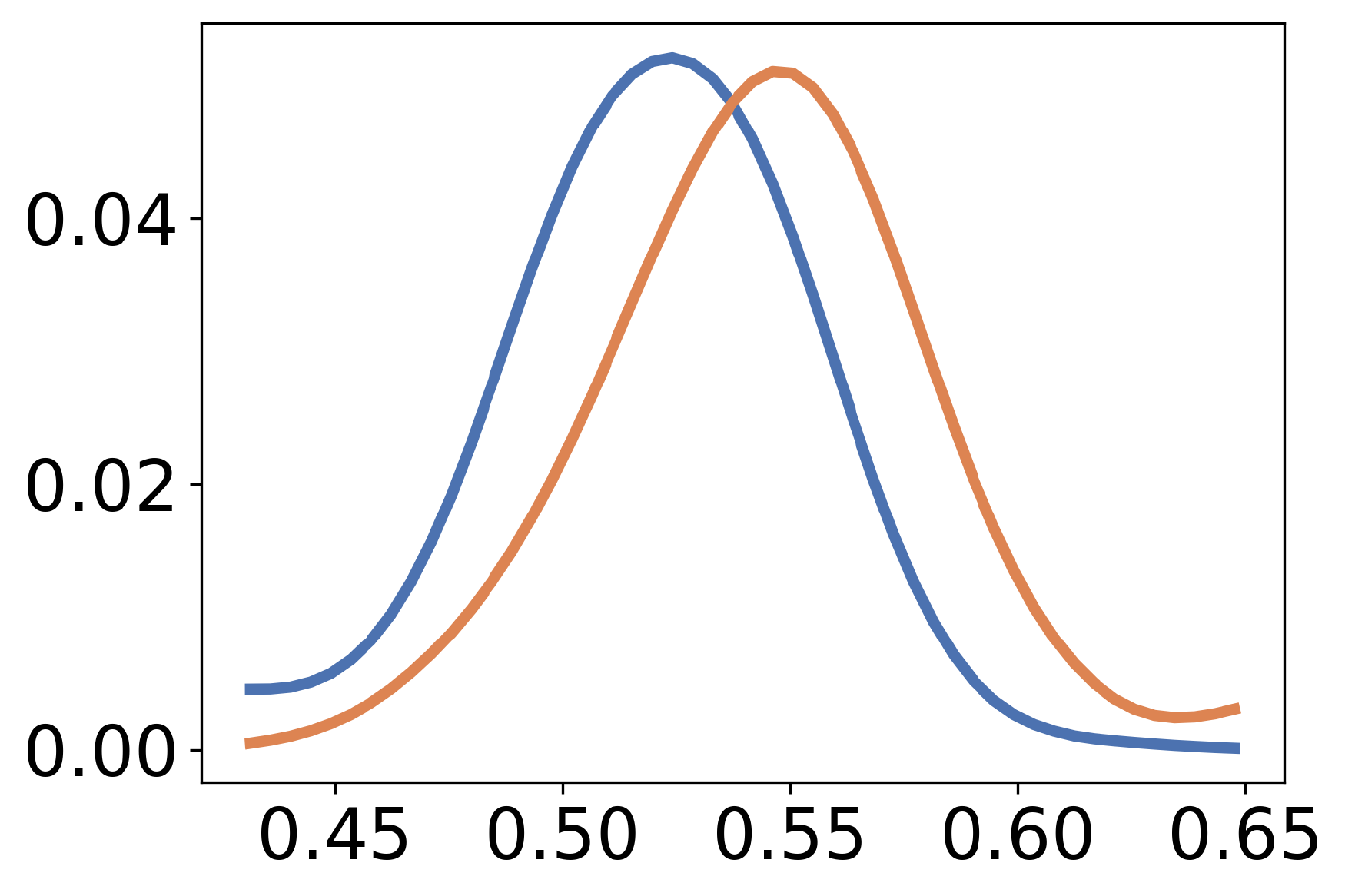}
        \caption{StanfordCars}
        \label{fig:stanford-cars}
    \end{subfigure}
    \hfill
    \begin{subfigure}[b]{0.22\textwidth}
        \includegraphics[width=\linewidth]{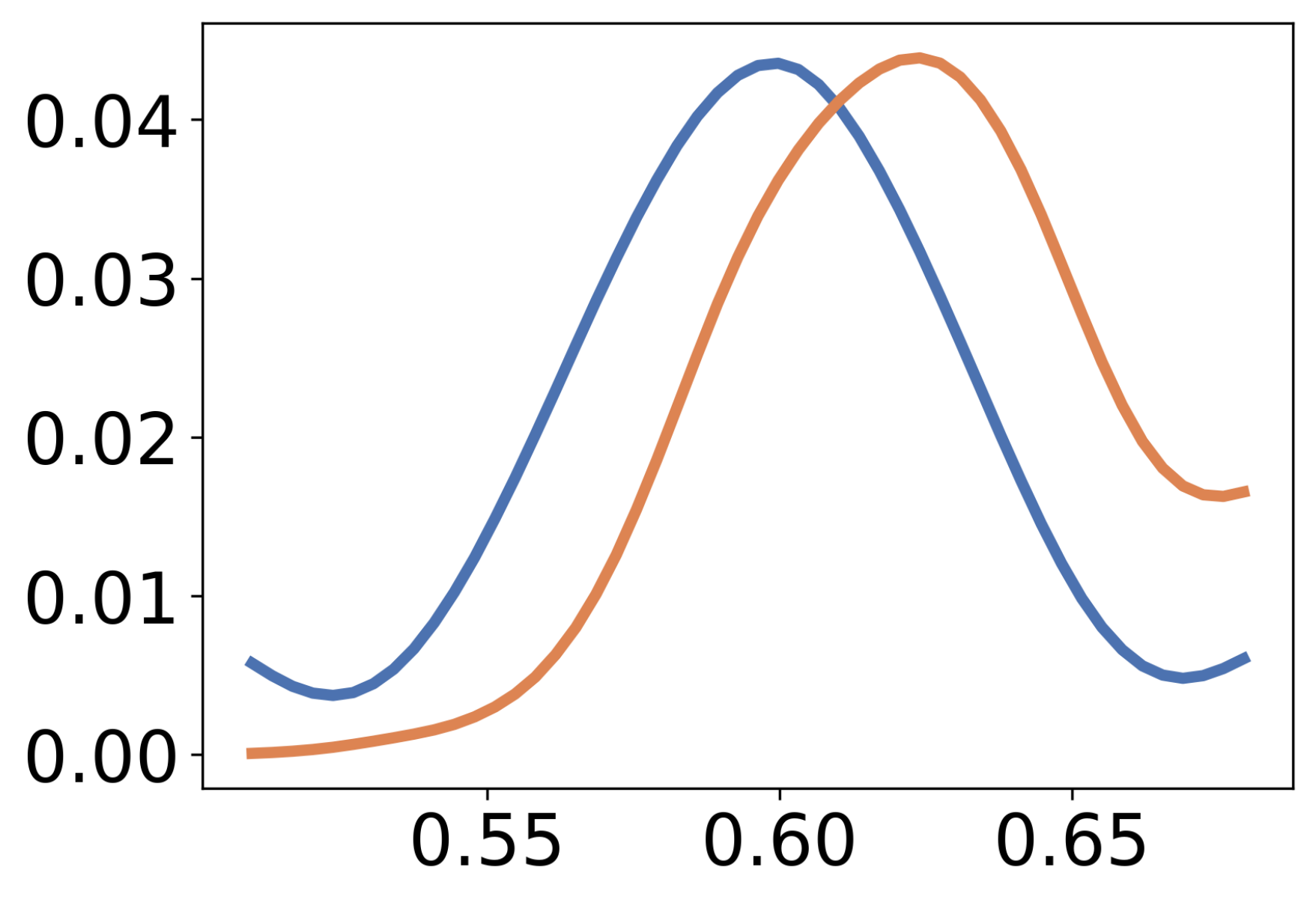}
        \caption{UCF-101}
        \label{fig:ucf-101}
    \end{subfigure}
    \caption{Distribution of cosine distances between text anchors and vision class-prototypes. TASSO (in {\color{mplBlue} blue}) consistently achieves lower distances than SnD (the closest competitor, in {\color{mplOrange} orange}). Metrics computed in the last continual step of $\mathbf{S^1}$ ordering.} %\fb{new version seems better, but is cropped to the top. Add back the x/y ticks with a bigger font; the legend can be removed - it's in the caption.}}
    \label{fig:cosine_dist}
\end{figure*}

%%%%%%%%%%%%%%%%%%%%%%%%%%

%%%%%%%%%%%%%%%%%%%%%%%%%%%%%%%%%

\begin{figure*}[t]
    \centering
    \includegraphics[width=\linewidth]{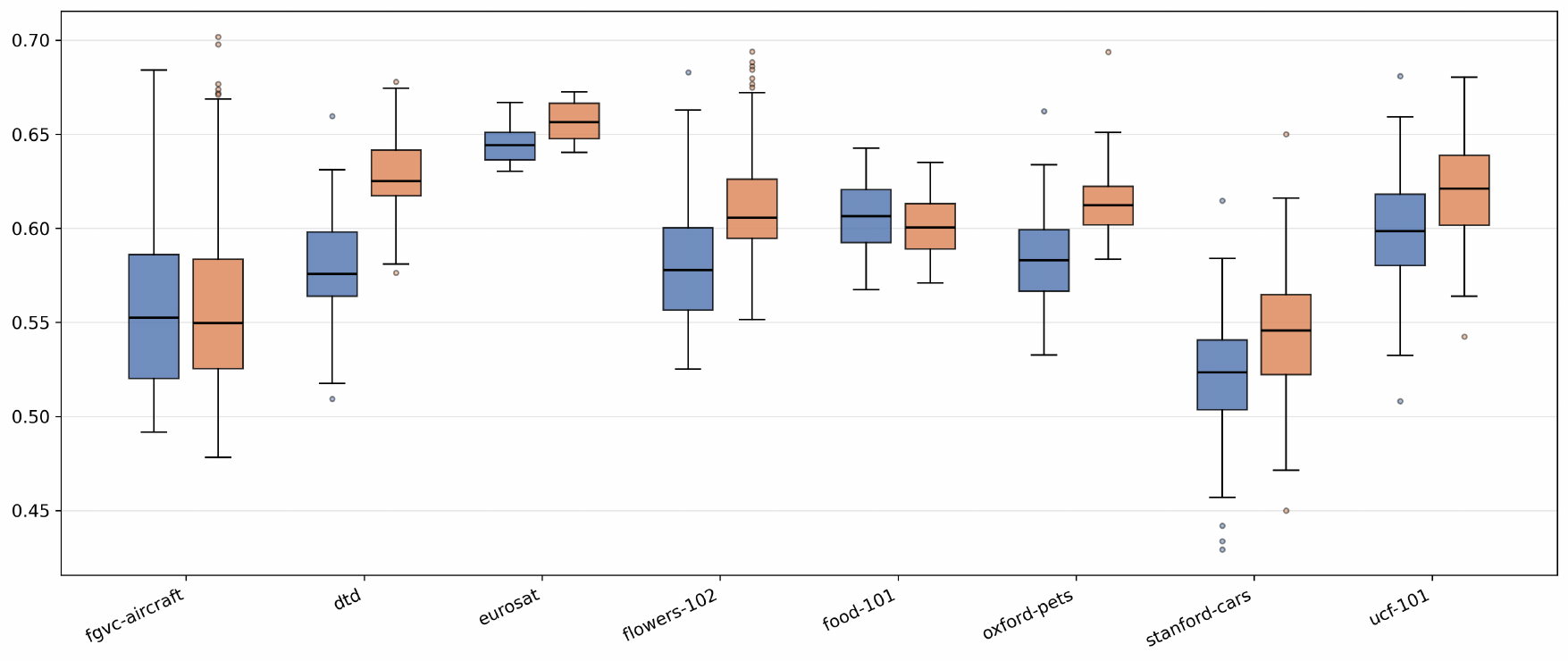}
    \caption{Stacked boxplot comparing the cosine distances between image prototypes and textual class embeddings (lower is better). TASSO (in {\color{mplBlue} blue}) and SnD (the closest competitor, in {\color{mplOrange} orange}). Metrics computed in the last continual step of $\mathbf{S^1}$ ordering.} %\fb{same as before, remove the legend which is in the caption already.}}
    \label{fig:boxplot}
\end{figure*}

As supporting experiments for the results reported in Tables \ref{tab:main_results_MTIL} and \ref{tab:main_results_MCIL}, in Figure \ref{fig:cosine_dist} we report the distributions of the distances between the text and vision class prototypes for TASSO and its closest competitor, SnD \cite{yu2024select}, confirming a significant improvement in multimodal alignment at the end of the Task-Incremental Learning process in 7 out of 8 tasks (only the Food-101 dataset shows a slight advantage of the competitor). 
A summarized version of the distributions is also shown in Figure \ref{fig:boxplot} as boxplots, confirming the analysis. Both figures refer to the $\mathbf{S^1}$ ordering. Overall, these experimental results confirm that TASSO is a simple yet effective approach to tackling Task-Incremental Learning of VLMs, surpassing the previous state-of-the-art by significant margins and setting new benchmark results in this field.

\subsection{Ablation Study}\label{sec:ablation_main}
Here, we provide supporting results for the design choices made in this work. %In particular, we report on a component analysis of both subspace learning and the knowledge distillation distance, as well as a hyperparameter value study. 
All ablation experiments are presented on sequence $\mathbf{S^1}$ in the MTIL setting.

\textbf{Ablation on the method components:}
we start with Table~\ref{tab:abl_comp}, where we analyze the impact of the two main contributions of this paper. 
The Table confirms that the joint use of our techniques yields the best results. % in terms of accuracy, forgetting, and zero-shot degradation.
As expected, the best performance improvements are achieved when the knowledge distillation objective is enabled. The standard L2 distance (recall that for normalized vectors like in our case, L2 is proportional to cosine distance, so the results for cosine distance are the same as L2) leads to remarkable results, but our geometry-aware geodesic metric achieves the best relative performance, with improvements of $0.6\%$ and $1\%$ in accuracy for the standard case and when using subspace learning, respectively. Note how the improvement is larger when combined with subspace learning, proving that the two contributions work well together and are complementary.
Furthermore, it allows for significant gains in catastrophic forgetting (which is, on average, halved) and even greater reductions in zero-shot degradation, which decreases by $\approx\!6$ times. 
In the \textit{Suppl. Mat.} we also show that computing the knowledge distillation separately on the two subspaces improves performances, specially in terms of zero-shot and catastrophic forgetting.
Regarding subspace learning, the Table clearly shows that it can improve accuracy, especially when combined with the geodesic metric, which further enhances performance by $0.7\%$, reaching the best score of $85.74\%$ and the best forgetting score of $0.81$, about $20\%$ better than the value of $1.00$ achieved with just the geodesic distance.  For zero-shot degradation, the biggest improvements come from the geodesic loss, but the combined use of both components leads to the optimal balance across the 3 metrics.
See the \textit{Suppl. Mat.} for an analysis of alternative strategies for the subspace optimization.
The ablation confirms the usefulness of both components in the efficient adaptation of VLM architectures and shows that they provide complementary contributions.

\begin{table}[t]
    \centering
    \caption{Ablation study on the method components in the MTIL setting}
    \label{tab:abl_comp}
    \begin{tabular}{c|c|c|c|c}
        \multirow{2}{*}{$\mathcal{L}_{KD}$} & Subspace & \multirow{2}{*}{Accuracy} & \multirow{2}{*}{Forgetting} & Zero-Shot \\
        & Learning & & & Degradation \\
        \hline
        \xmark & \xmark  & 76.86 & 9.82 & 11.63 \\ % server033
        $L2$ & \xmark  & 84.48 & 1.71 & 1.75 \\ % server086
        %$\cos^{-1}\left<\cdot,\cdot\right>$ & \xmark  & &  &  \\ 
        $\mathcal{L}_{\text{geo}}$ & \xmark & 85.07 & 1.00 & \textbf{0.25}\\% server004
        \hdashline
        \xmark & \cmark  & 75.24 & 13.01 & 17.60 \\ % server173
        $L2$ & \cmark  & 84.73 & 2.26 & 1.86 \\ % server070
        %$\cos^{-1}\left<\cdot,\cdot\right>$ & \cmark  & 85.74 & 0.81 & 0.36 \\
        % full embedding version
        %$\mathcal{L}_{\text{geo}+\text{full}}$ & \cmark & 85.72 & 1.21 & 0.59 \\
        $\mathcal{L}_{\text{geo}}$ & \cmark  & \textbf{85.74} & \textbf{0.81} & 0.36 \\
        
    \end{tabular}
\end{table}

\textbf{Ablation on the hyperparameters:}
we continue the ablation studies in Table~\ref{tab:abl_hyper}, where we analyze the impact of the main hyperparameters of the method: the rank $r$ of the projector matrix $U$, the subspace learning loss scale $\alpha$, and the knowledge distillation loss scale $\beta$.
For all hyperparameters, we report three values: our best choice, a smaller setting, and a larger one. In all cases, we observe the relative stability of the metrics across the various configurations, indicating that parameter sensitivity is relatively low in our architecture.
More specifically, for the rank parameter $r$, the selected value of $144$ leads to the best performance on all 3 metrics, even though halving or doubling the dimension results in a similar accuracy.
For the subspace learning loss weight, the accuracy tends to increase with the value of $\alpha$, but forgetting and zero-shot degradation worsen with larger values. The selected value of $0.5$ achieves a good trade-off across all 3 metrics.
Finally, setting the knowledge distillation loss weight $\beta=3$ leads to almost optimal accuracy, with very good forgetting and zero-shot scores. The smaller value of $1.5$ results in an almost unnoticeable improvement in accuracy at the cost of much worse forgetting and zero-shot performances.

\noindent
Additional ablation studies and the pseudocode  are provided in the \textit{Suppl. Mat.}

\begin{table}[t]
    \caption{Ablation study on the method hyperparameters in the MTIL sequence $\mathbf{S^1}$ setting: a) the rank $r$ of the projector matrix $U$; b) subspace learning loss scale $\alpha$; c) knowledge distillation loss weight $\beta$.}
    \label{tab:abl_hyper}
    \centering
\begin{minipage}{0.3\textwidth}
    \centering
    \begin{tabular}{c|c|c|c}
        $r$    & Acc. & Forg. & Z.Shot \\
        \hline
        72 & 85.70 & 0.85 & 0.41\\ % server033
        %128 & 85.79 & 0.88 & 0.39 \\ % server070
        \textbf{144}  &85.74 & 0.81 & 0.36  \\
        %192 & 85.67 & 0.76 & 0.36\\ % server 173
        288 & 85.72 & 0.85 & 0.38\\ % server004
    \end{tabular} 
    \centering

    a)
    \end{minipage}
    \begin{minipage}{0.3\textwidth}
    \centering
    \begin{tabular}{c|c|c|c}
        $\alpha$     & Acc. & Forg. & Z.Shot \\
        \hline
        0.1 & 85.19 & 0.76 & 0.24\\ % server070
        %0.2 & 85.40 & 0.79 & 0.26 \\ 
        \textbf{0.5} & 85.74 & 0.81 & 0.36  \\
        %0.8 & 85.84 & 0.76 & 0.50\\
        1 & 85.86 & 0.99 & 0.57\\ % server173
    \end{tabular} 
    \centering

    b)
    \end{minipage}
    \begin{minipage}{0.3\textwidth}
    \centering
    \begin{tabular}{c|c|c|c}
        $\beta$  & Acc. & Forg. & Z.Shot \\
        \hline
        1.5  & 85.77 & 1.11 & 0.61\\ 
        \textbf{3} & 85.74 & 0.81 & 0.36  \\
        %4.5  & 85.64 & 0.50 & 0.34 \\
        6 & 85.37 & 0.68 & 0.22 \\ % server095
    \end{tabular} 
    \centering

    c)
    \end{minipage}
 \end{table}

%\subsubsection{Ablation on the subspace learning distances}
%In Table~\ref{tab:xxx} we analyze ... The Table shows that....

\section{Conclusions and Future Work}
\label{sec:conclusion}
In this paper, we present TASSO (TAsk-Specific Subspace Optimization), a novel approach to Task Incremental Learning for Vision-Language Models. It introduces a subspace learning strategy and a modified distance metric for distillation that allows it to outperform previous state-of-the-art techniques by more than $1\%$ in accuracy on both Multidomain Task Incremental Learning (MTIL) and Multidomain Class Incremental Learning (MCIL) benchmarks. 
Even more remarkable gains are made in Catastrophic Forgetting and Zero-Shot Degradation, both of which are drastically reduced in the two settings. Unlike some competitors, our approach can tackle the task using a single teacher model and does not require computationally demanding additional modules or stages, thanks to the combination of subspace learning and geometry-aware knowledge distillation.

The results in the paper highlight the effectiveness of subspace learning, and future research directions will further explore this topic. More specifically, we plan to implement dynamic low-rank approximation for different tasks, allowing for an adaptive setting of the dimensionality of the subspace based on the semantics of the domains. Moreover, text-image alignment is also another important direction as this paper only adapts the visual side.
We also plan to explore the application of the approach to different VLMs and to other computer vision tasks beyond image classification to further verify the generalizability of the method.

%%%%%%%%%%%%%%%%%%%%%%%%%%%%%%%%%%%%%%%%%%%%%%%%%%%%%%%%%%%%%%%%%%%%%%%%%%%%%%%%%%%%%%%%%%%%%%%%%%%
% \section*{Acknowledgments}

\clearpage

\bibliographystyle{splncs04}
\bibliography{main}

@String(PAMI  = {IEEE Trans. Pattern Anal. Mach. Intell.})

@String(CVPR  = {IEEE Conf. Comput. Vis. Pattern Recog.})

@String(ICCV  = {Int. Conf. Comput. Vis.})

@String(ECCV  = {Eur. Conf. Comput. Vis.})

@String(NeurIPS = {Adv. Neural Inform. Process. Syst.})

@String(ICML  = {Int. Conf. Mach. Learn.})

@String(ICLR  = {Int. Conf. Learn. Represent.})

@inproceedings{radford2021learning,
  title={Learning transferable visual models from natural language supervision},
  author={Radford, Alec and Kim, Jong Wook and Hallacy, Chris and Ramesh, Aditya and Goh, Gabriel and Agarwal, Sandhini and Sastry, Girish and Askell, Amanda and Mishkin, Pamela and Clark, Jack and others},
  booktitle=ICML,
  pages={8748--8763},
  year={2021},
  organization={PMLR}
}

@inproceedings{li2022blip,
  title={Blip: Bootstrapping language-image pre-training for unified vision-language understanding and generation},
  author={Li, Junnan and Li, Dongxu and Xiong, Caiming and Hoi, Steven},
  booktitle=ICML,
  pages={12888--12900},
  year={2022},
  organization={PMLR}
}

@inproceedings{jia2021scaling,
  title={Scaling up visual and vision-language representation learning with noisy text supervision},
  author={Jia, Chao and Yang, Yinfei and Xia, Ye and Chen, Yi-Ting and Parekh, Zarana and Pham, Hieu and Le, Quoc and Sung, Yun-Hsuan and Li, Zhen and Duerig, Tom},
  booktitle=ICML,
  pages={4904--4916},
  year={2021},
  organization={PMLR}
}

@article{alayrac2022flamingo,
  title={Flamingo: a visual language model for few-shot learning},
  author={Alayrac, Jean-Baptiste and Donahue, Jeff and Luc, Pauline and Miech, Antoine and Barr, Iain and Hasson, Yana and Lenc, Karel and Mensch, Arthur and Millican, Katherine and Reynolds, Malcolm and others},
  journal=NeurIPS,
  volume={35},
  pages={23716--23736},
  year={2022}
}

@article{liu2025continual,
  title={Continual learning for VLMs: A survey and taxonomy beyond forgetting},
  author={Liu, Yuyang and Hong, Qiuhe and Huang, Linlan and Gomez-Villa, Alexandra and Goswami, Dipam and Liu, Xialei and van de Weijer, Joost and Tian, Yonghong},
  journal={arXiv preprint arXiv:2508.04227},
  year={2025}
}

@article{van2019three,
  title={Three scenarios for continual learning},
  author={Van de Ven, Gido M and Tolias, Andreas S},
  journal={arXiv preprint arXiv:1904.07734},
  year={2019}
}

@inproceedings{zheng2023preventing,
  title={Preventing zero-shot transfer degradation in continual learning of vision-language models},
  author={Zheng, Zangwei and Ma, Mingyuan and Wang, Kai and Qin, Ziheng and Yue, Xiangyu and You, Yang},
  booktitle=ICCV,
  pages={19125--19136},
  year={2023}
}

@inproceedings{yu2024select,
  title={Select and distill: Selective dual-teacher knowledge transfer for continual learning on vision-language models},
  author={Yu, Yu-Chu and Huang, Chi-Pin and Chen, Jr-Jen and Chang, Kai-Po and Lai, Yung-Hsuan and Yang, Fu-En and Wang, Yu-Chiang Frank},
  booktitle=ECCV,
  pages={219--236},
  year={2024},
  organization={Springer}
}

@inproceedings{zheng2024adapt,
  title={Adapt without forgetting: Distill proximity from dual teachers in vision-language models},
  author={Zheng, Mengyu and Tang, Yehui and Hao, Zhiwei and Han, Kai and Wang, Yunhe and Xu, Chang},
  booktitle=ECCV,
  pages={109--125},
  year={2024},
  organization={Springer}
}

@article{ilharco2021openclip,
  title={Openclip},
  author={Ilharco, Gabriel and Wortsman, Mitchell and Carlini, Nicholas and Taori, Rohan and Dave, Achal and Shankar, Vaishaal and Namkoong, Hongseok and Miller, John and Hajishirzi, Hannaneh and Farhadi, Ali and others},
  journal={Zenodo},
  year={2021}
}

@article{dosovitskiy2020image,
  title={An image is worth 16x16 words: Transformers for image recognition at scale},
  author={Dosovitskiy, Alexey and Beyer, Lucas and Kolesnikov, Alexander and Weissenborn, Dirk and Zhai, Xiaohua and Unterthiner, Thomas and Dehghani, Mostafa and Minderer, Matthias and Heigold, Georg and Gelly, Sylvain and others},
  journal={arXiv preprint arXiv:2010.11929},
  year={2020}
}

@article{maji2013fine,
  title={Fine-grained visual classification of aircraft},
  author={Maji, Subhransu and Rahtu, Esa and Kannala, Juho and Blaschko, Matthew and Vedaldi, Andrea},
  journal={arXiv preprint arXiv:1306.5151},
  year={2013}
}

@inproceedings{cimpoi2014describing,
  title={Describing textures in the wild},
  author={Cimpoi, Mircea and Maji, Subhransu and Kokkinos, Iasonas and Mohamed, Sammy and Vedaldi, Andrea},
  booktitle=CVPR,
  pages={3606--3613},
  year={2014}
}

@article{helber2019eurosat,
  title={Eurosat: A novel dataset and deep learning benchmark for land use and land cover classification},
  author={Helber, Patrick and Bischke, Benjamin and Dengel, Andreas and Borth, Damian},
  journal={IEEE Journal of Selected Topics in Applied Earth Observations and Remote Sensing},
  volume={12},
  number={7},
  pages={2217--2226},
  year={2019},
  publisher={IEEE}
}

@inproceedings{nilsback2008automated,
  title={Automated flower classification over a large number of classes},
  author={Nilsback, Maria-Elena and Zisserman, Andrew},
  booktitle={2008 Sixth Indian conference on computer vision, graphics \& image processing},
  pages={722--729},
  year={2008},
  organization={IEEE}
}

@inproceedings{bossard2014food,
  title={Food-101--mining discriminative components with random forests},
  author={Bossard, Lukas and Guillaumin, Matthieu and Van Gool, Luc},
  booktitle=ECCV,
  pages={446--461},
  year={2014},
  organization={Springer}
}

@inproceedings{parkhi2012cats,
  title={Cats and dogs},
  author={Parkhi, Omkar M and Vedaldi, Andrea and Zisserman, Andrew and Jawahar, CV},
  booktitle=CVPR,
  pages={3498--3505},
  year={2012},
  organization={IEEE}
}

@inproceedings{krause20133d,
  title={3d object representations for fine-grained categorization},
  author={Krause, Jonathan and Stark, Michael and Deng, Jia and Fei-Fei, Li},
  booktitle={Proceedings of the IEEE international conference on computer vision workshops},
  pages={554--561},
  year={2013}
}

@article{khurram2012dataset,
  title={A dataset of 101 human action classes from videos in the wild},
  author={Khurram, Soomro and Roshan, Zamir Amir and Shah, M},
  journal={Center for Research in Computer Vision},
  volume={2},
  number={11},
  year={2012}
}

@article{bell2022effect,
  title={The effect of task ordering in continual learning},
  author={Bell, Samuel J and Lawrence, Neil D},
  journal={arXiv preprint arXiv:2205.13323},
  year={2022}
}

@article{li2025optimal,
  title={Optimal task order for continual learning of multiple tasks},
  author={Li, Ziyan and Hiratani, Naoki},
  journal={arXiv preprint arXiv:2502.03350},
  year={2025}
}

@inproceedings{chaudhry2018riemannian,
  title={Riemannian walk for incremental learning: Understanding forgetting and intransigence},
  author={Chaudhry, Arslan and Dokania, Puneet K and Ajanthan, Thalaiyasingam and Torr, Philip HS},
  booktitle=ECCV,
  pages={532--547},
  year={2018}
}

@article{chaudhry2019tiny,
  title={On tiny episodic memories in continual learning},
  author={Chaudhry, Arslan and Rohrbach, Marcus and Elhoseiny, Mohamed and Ajanthan, Thalaiyasingam and Dokania, Puneet K and Torr, Philip HS and Ranzato, Marc'Aurelio},
  journal={arXiv preprint arXiv:1902.10486},
  year={2019}
}

@article{lopez2017gradient,
  title={Gradient episodic memory for continual learning},
  author={Lopez-Paz, David and Ranzato, Marc'Aurelio},
  journal=NeurIPS,
  volume={30},
  year={2017}
}

@article{li2017learning,
  title={Learning without forgetting},
  author={Li, Zhizhong and Hoiem, Derek},
  journal=PAMI,
  volume={40},
  number={12},
  pages={2935--2947},
  year={2017},
  publisher={IEEE}
}

@inproceedings{rebuffi2017icarl,
  title={icarl: Incremental classifier and representation learning},
  author={Rebuffi, Sylvestre-Alvise and Kolesnikov, Alexander and Sperl, Georg and Lampert, Christoph H},
  booktitle=CVPR,
  pages={2001--2010},
  year={2017}
}

@inproceedings{yu2024boosting,
  title={Boosting continual learning of vision-language models via mixture-of-experts adapters},
  author={Yu, Jiazuo and Zhuge, Yunzhi and Zhang, Lu and Hu, Ping and Wang, Dong and Lu, Huchuan and He, You},
  booktitle=CVPR,
  pages={23219--23230},
  year={2024}
}

@article{hu2022lora,
  title={Lora: Low-rank adaptation of large language models.},
  author={Hu, Edward J and Shen, Yelong and Wallis, Phillip and Allen-Zhu, Zeyuan and Li, Yuanzhi and Wang, Shean and Wang, Liang and Chen, Weizhu and others},
  journal=ICLR,
  volume={1},
  number={2},
  pages={3},
  year={2022}
}

@inproceedings{houlsby2019parameter,
  title={Parameter-efficient transfer learning for NLP},
  author={Houlsby, Neil and Giurgiu, Andrei and Jastrzebski, Stanislaw and Morrone, Bruna and De Laroussilhe, Quentin and Gesmundo, Andrea and Attariyan, Mona and Gelly, Sylvain},
  booktitle=ICML,
  pages={2790--2799},
  year={2019},
  organization={PMLR}
}

@inproceedings{wang2021k,
  title={K-adapter: Infusing knowledge into pre-trained models with adapters},
  author={Wang, Ruize and Tang, Duyu and Duan, Nan and Wei, Zhongyu and Huang, Xuan-Jing and Ji, Jianshu and Cao, Guihong and Jiang, Daxin and Zhou, Ming},
  booktitle={Findings of the Association for Computational Linguistics: ACL-IJCNLP 2021},
  pages={1405--1418},
  year={2021}
}

@article{lu2024adaptive,
  title={Adaptive rank, reduced forgetting: Knowledge retention in continual learning vision-language models with dynamic rank-selective lora},
  author={Lu, Haodong and Zhao, Chongyang and Xue, Jason and Yao, Lina and Moore, Kristen and Gong, Dong},
  journal={arXiv preprint arXiv:2412.01004},
  year={2024}
}

@article{jacobs1991adaptive,
  title={Adaptive mixtures of local experts},
  author={Jacobs, Robert A and Jordan, Michael I and Nowlan, Steven J and Hinton, Geoffrey E},
  journal={Neural computation},
  volume={3},
  number={1},
  pages={79--87},
  year={1991},
  publisher={MIT Press}
}

@inproceedings{kang2025dynamic,
  title={Dynamic multi-layer null space projection for vision-language continual learning},
  author={Kang, Borui and Wang, Lei and Wu, Zhiping and Feng, Tao and Li, Yawen and Gao, Yang and Li, Wenbin},
  booktitle=ICCV,
  pages={2077--2086},
  year={2025}
}

@article{li2024continual,
  title={Continual learning with knowledge distillation: A survey},
  author={Li, Songze and Su, Tonghua and Zhang, Xu-Yao and Wang, Zhongjie},
  journal={IEEE Transactions on Neural Networks and Learning Systems},
  volume={36},
  number={6},
  pages={9798--9818},
  year={2024},
  publisher={IEEE}
}

@article{wang2022s,
  title={S-prompts learning with pre-trained transformers: An occam’s razor for domain incremental learning},
  author={Wang, Yabin and Huang, Zhiwu and Hong, Xiaopeng},
  journal=NeurIPS,
  volume={35},
  pages={5682--5695},
  year={2022}
}

@article{li2025coleclip,
  title={Coleclip: Open-domain continual learning via joint task prompt and vocabulary learning},
  author={Li, Yukun and Pang, Guansong and Suo, Wei and Jing, Chenchen and Xi, Yuling and Liu, Lingqiao and Chen, Hao and Liang, Guoqiang and Wang, Peng},
  journal={IEEE Transactions on Neural Networks and Learning Systems},
  year={2025},
  publisher={IEEE}
}

@inproceedings{liu2025c,
  title={C-CLIP: Multimodal continual learning for vision-language model},
  author={Liu, Wenzhuo and Zhu, Fei and Wei, Longhui and Tian, Qi},
  booktitle=ICLR,
  year={2025}
}

@article{garg2023tic,
  title={Tic-clip: Continual training of clip models},
  author={Garg, Saurabh and Farajtabar, Mehrdad and Pouransari, Hadi and Vemulapalli, Raviteja and Mehta, Sachin and Tuzel, Oncel and Shankar, Vaishaal and Faghri, Fartash},
  journal={arXiv preprint arXiv:2310.16226},
  year={2023}
}

@article{wang2024comprehensive,
  title={A comprehensive survey of continual learning: Theory, method and application},
  author={Wang, Liyuan and Zhang, Xingxing and Su, Hang and Zhu, Jun},
  journal=PAMI,
  volume={46},
  number={8},
  pages={5362--5383},
  year={2024},
  publisher={IEEE}
}

@misc{mei2025geommgeodesicperspectivemultimodal,
      title={GeoMM: On Geodesic Perspective for Multi-modal Learning}, 
      author={Shibin Mei and Hang Wang and Bingbing Ni},
      year={2025},
      eprint={2505.11216},
      archivePrefix={arXiv},
      primaryClass={cs.CV},
      _url={https://arxiv.org/abs/2505.11216}, 
}

@inproceedings{kang2025clip,
  title={Is CLIP ideal? No. Can we fix it? Yes!},
  author={Kang, Raphi and Song, Yue and Gkioxari, Georgia and Perona, Pietro},
  booktitle=ICCV,
  pages={22436--22446},
  year={2025}
}

@inproceedings{deng2009imagenet,
  author    = {Jia Deng and
               Wei Dong and
               Richard Socher and
               Li{-}Jia Li and
               Kai Li and
               Fei{-}Fei Li},
  title     = {ImageNet: {A} large-scale hierarchical image database},
  pages     = {248--255},
  booktitle = CVPR,
  year      = {2009}
}

@article{wu2025synthetic,
  _title={Synthetic data is an elegant gift for continual vision-language models},
  title={Synth. data is an elegant gift for cont. {VLMs}},
  __title={{GIFT}},
  _author={Wu, Bin and Shi, Wuxuan and Wang, Jinqiao and Ye, Mang},
  author={Wu and others},
  _booktitle={CVPR},
  journal={CVPR},
  _booktitle=CVPR,
  _pages={2813--2823},
  year={2025}
}

\clearpage

\appendix
\section{Appendix Summary}
    In this document, we present supplementary experiments and clarifications that allow us to better motivate the design choices behind the TASSO approach and to evaluate its performance in more detail. 
Specifically, in Section \ref{sec:ablation} we report additional ablation studies examining various facets of our continual learning strategy. 
In Section \ref{sec:compute} we report the computational overhead related to the proposed approach.
We then move to Section \ref{sec:confusion}, where we include the per-domain results at each incremental training step associated with the summary metrics reported in the main document for sequence $\mathbf{S^1}$. 
Furthermore, in Section \ref{sec:stats}, we conduct a statistical analysis of the results presented in the main paper. 
Next,  we describe in more detail the datasets and task sequences employed in our experimental evaluation in Section \ref{sec:datasets}.
Finally, in Section \ref{sec:pseudocode} we provide the pseudocode for our method, including training and evaluation procedures for the Multidomain Task (MTIL) and Class (MCIL) Incremental settings.

\section{Additional Ablation Studies}\label{sec:ablation}
In this section, we present additional ablation studies that encompass all aspects of the TASSO multidomain learning approach.
We start by reporting the results of the components and hyperparameter ablation studies on the MCIL benchmark in Sec.~\ref{subsec:abl_mcil} and Sec.~\ref{subsec:hyper_mcil}, respectively. 
This complements the study in the main document, which reported results only on the MTIL benchmark.
We also report in Sec.~\ref{subsec:decomposition} the advantages of decomposing embeddings in knowledge distillation computation. %compared to using the full embedding. 
In Sec.~\ref{subsec:optimization} we fully analyze the choice of jointly optimizing both the image encoder and the subspace in our subspace learning loss $\mathcal{L}_{sub}$.
We then analyze the performance of our approach when employing different Knowledge Distillation (KD) distance functions in the two subspaces identified by TASSO (namely, task-specific and task-irrelevant) in Sec.~\ref{subsec:kd}.

% \begin{table}[t]
% \caption{Ablation study on the method hyperparameters in the MCIL sequence $\mathbf{S^1}$ setting: a) the rank $r$ of the projector matrix $U$; b) subspace learning loss scale $\alpha$; c) knowledge distillation loss weight $\beta$.}
% \label{tab:abl_hyper}
% \centering
% \begin{minipage}{0.3\textwidth}
% \centering
% \begin{tabular}{c|c|c|c}
%     $r$    & Acc. & Forg. & Z.Shot \\
%     \hline
%     72 & 84.99 & 1.02 & 0.54 \\ % server 107
%     144 & 85.00 & 1.05 & 0.47 \\
%     288 & 84.23 & 0.99 & 0.50\\ % server 070
% \end{tabular} 
% \centering

% a)
% \end{minipage}
% \begin{minipage}{0.3\textwidth}
% \centering
% \begin{tabular}{c|c|c|c}
%     $\alpha$     & Acc. & Forg. & Z.Shot \\
%     \hline
%     0.1 & 84.16 & 0.87 & 0.29\\
%     0.5 & 85.00 & 1.05 & 0.47\\
%     1 & 85.08 & 1.40 & 0.90 \\
% \end{tabular} 
% \centering

% b)
% \end{minipage}
% \begin{minipage}{0.3\textwidth}
% \centering
% \begin{tabular}{c|c|c|c}
%     $\beta$  & Acc. & Forg. & Z.Shot \\
%     \hline
%     1.5 & 85.14 & 1.28 & 0.84\\ % server070
%     3 & 85.00 & 1.05 & 0.47 \\
%     6 & 84.59 & 0.80 &0.34 \\ % server173
% \end{tabular} 
% \centering

% c)
% \end{minipage}
% \end{table}

\subsection{Ablation on the method's components on MCIL}
\label{subsec:abl_mcil}
In this section, we report the results of the ablation study on the various method components on the MCIL benchmark (in the main document, the study was performed only on the MTIL benchmark).
More specifically, we report TASSO's accuracy in the MCIL task using the same ablated setting presented in Table 3 of the main paper.

The results are reported in Tab. \ref{tab:abl_main} and closely match those shown in the main document. More specifically, the best performance is achieved when the knowledge distillation objective is combined with subspace learning. Geodesic distillation is consistently better than L2 with or without subspace learning. When subspace learning is enabled, geodesic distillation reduces the forgetting and zero-shot degradation by $1.72\%$ and $2.32\%$, respectively. Even without subspace learning, geodesic distillation is still able to mitigate both the forgetting and zero-shot degradation by $1.80\%$ and $1.33\%$. This indicates that the main preservation effect comes from the distillation term, and that matching geometry on the hypersphere is more effective than Eculidean alignment.
The subspace learning strategy allows for a further boost in accuracy of $1.17\%$ reaching the best value of $85.00\%$ when both strategies are enabled.

\begin{table}[t]
    \centering
    \caption{Ablation study on the method components in the MCIL setting.}
    \label{tab:abl_main}
    \begin{tabular}{c|c|c|c|c}
        \multirow{2}{*}{$\mathcal{L}_{KD}$} & Subspace & \multirow{2}{*}{Accuracy} & \multirow{2}{*}{Forgetting} & Zero-Shot \\
        & Learning & & & Degradation \\
        \hline
        \xmark & \xmark  & 75.70 & 9.99 & 13.73 \\ % server173
        $L2$ & \xmark  & 83.27 & 1.85 & 1.72\\ % server004
         
        $\mathcal{L}_{\text{geo}}$ & \xmark & 83.83 & \textbf{1.05} & \textbf{0.39} \\% server202
        \hdashline
        \xmark & \cmark  & 74.33 & 13.73 & 19.60 \\ % server173
        $L2$ & \cmark  & 83.83 & 2.77 & 2.79 \\ % server004
        $\mathcal{L}_{\text{geo}}$ & \cmark  & \textbf{85.00}  & \textbf{1.05}& 0.47\\
    \end{tabular}
\end{table}

\subsection{Hyperparameter Study on MCIL}\label{subsec:hyper_mcil}
In this section, we report the results of the ablation study on the MCIL benchmark (in the main document, the study was performed only on the MTIL benchmark). More specifically, we report TASSO's accuracy in the MCIL task as we vary its hyperparameters.

The results are reported in Tab. \ref{tab:abl_hyper_supple} and closely match those shown in the main document. More specifically, the selected subspace rank leads to the best accuracy in this task (Tab. \ref{subtab:hyper_rank}), while the chosen values of $\alpha$ and $\beta$ result in a very good tradeoff between metrics (with the second-best accuracy very close to the best, as well as small forgetting and zero-shot degradation performances; see Tab. \ref{subtab:hyper_alpha} and \ref{subtab:hyper_beta}). Moreover, note how the metrics across different hyperparameters' values   are quite constant, confirming that TASSO is hyperparameter-stable.

\begin{table}[t]
    \centering
    \caption{Ablation study on the hyperparameters. MCIL sequence $\mathbf{S^1}$.}
    \label{tab:abl_hyper_supple}
    \begin{subtable}{.3\linewidth}
        \centering
        \begin{tabular}{c|ccc}
            $r$ & Acc. & Forg. & Z.S. Deg. \\
            \hline
            72 & 84.99 & 1.02 & 0.54 \\
            144 & 85.00 & 1.05 & 0.47 \\
            288 & 84.23 & 0.99 & 0.50 \\
        \end{tabular}
        \caption{Supspace Rank}\label{subtab:hyper_rank}
    \end{subtable}
    \begin{subtable}{.3\linewidth}
        \centering
        \begin{tabular}{c|ccc}
            $\alpha$ & Acc. & Forg. & Z.S. Deg. \\
            \hline
            0.1 & 84.16 & 0.87 & 0.29 \\
            0.5 & 85.00 & 1.05 & 0.47 \\
            1.0 & 85.08 & 1.40 & 0.90 \\
        \end{tabular}
        \caption{$\mathcal{L}_{\text{sub}}$ coefficient.}\label{subtab:hyper_alpha}
    \end{subtable}
    \begin{subtable}{.3\linewidth}
        \centering
        \begin{tabular}{c|ccc}
            $\beta$ & Acc. & Forg. & Z.S. Deg. \\
            \hline
            1.5 & 85.14 & 1.28 & 0.84 \\
            3.0 & 85.00 & 1.05 & 0.47 \\
            6.0 & 84.59 & 0.80 & 0.34 \\
        \end{tabular}
        \caption{$\mathcal{L}_{\text{KD}}$ coefficient.}\label{subtab:hyper_beta}
    \end{subtable}
\end{table}

\subsection{Ablation on embedding decomposition}\label{subsec:decomposition}
We run an additional ablation study where the KD loss is applied on the original vectors, rather than on the subspace decomposition. 
The evaluation on the MTIL task results in an accuracy of $85.72$, a catastrophic forrgetting of $1.21$, and a zero-shot of $0.59$, that is, an almost identical accuracy, but significant degradation in CF and ZS ($33\%$ and $39\%$, respectively).

\subsection{Analysis of Subspace-Encoder Optimization Strategies}\label{subsec:optimization}
The subspace learning loss $\mathcal{L}_{sub}$ is meant to target both the image encoder and the projector. It encourages the image encoder to focus on a low-dimensional subspace while training the task-specific projector. Note that the subspace receives supervision from the text-encoder as well, which, being frozen, provides a clearer signal. As evidence for the joint optimization, in Tab.~\ref{tab:optimization}, we run an experiment where the image encoder does not receive supervision from $\mathcal{L}_{sub}$, \ie, when a stop-grad layer is added after the features. We denote it as TASSO (no $\mathcal{L}_{sub}$ sup.) and the results are: Acc $84.88$, CF $0.79$, ZS $0.13$; that is, a $1\%$ drop in accuracy. 

Furthermore, we also run two experiments where the subspace and the encoder are optimized separately. TASSO (proj-dec) optimizes the projector first and the decoder second, achieving Acc $85.37$, CF $1.21$, ZS $0.44$, \ie, a reduction in all metrics. By alternating the training of the two components (odd step projector, even encoder), TASSO (alt) achieves Acc $85.51$, CF $0.70$, and ZS $0.20$ (\ie, slightly worse Acc, but better CF and ZS). Summarizing, joint training achieves the best results, but alternating training as suggested is a viable option.

% \pz[]{here I would place a table}
\begin{table}[]
    \centering
    \caption{Different optimization choices for the subspace learning loss $\mathcal{L}_{sub}$.}
    \begin{tabular}{c|c|c|c}
    \toprule
    & Accuracy & Forgetting & Z.S. Degradation \\
    \hline
    TASSO & 85.74 & 0.81 & 0.36 \\
    \hdashline
    TASSO(no $\mathcal{L}_{sub}$ sup.) & 84.88 & 0.79 & 0.13 \\
    TASSO (proj-dec) & 85.37 & 1.21 & 0.44 \\
    TASSO (alt) & 85.51 & 0.70 & 0.20 \\
    \bottomrule
    \end{tabular}
    \label{tab:optimization}
\end{table}

\subsection{Knowledge Distillation}\label{subsec:kd}

In this section, we evaluate the effect of using different knowledge distillation techniques in the two subspaces identified by TASSO. More specifically, we test three configurations for each subspace: i) No KD (only subspace learning is employed), ii) L2 distance (which is also proportional to the cosine distance for normalized vectors), and iii) our proposed geodesic distance.
The results reported in Tables \ref{subtab:kd_acc}-\ref{subtab:kd_deg} confirm the optimality of our distillation objective, which achieves second-best accuracy (very close to the best Geo/L2 configuration) and the best forgetting and zero-shot degradation results.

\begin{table}[h]
    \centering
    \caption{Subspace distillation ablation}
    \begin{subtable}{.3\linewidth}
        \centering
        \begin{tabular}{c|ccc}
            \diagbox{$\mathbf{f}_\perp$}{$\mathbf{f}_\parallel$} & \xmark & L2 & Geo \\
            \hline
            \xmark & 75.24 & 83.05 & 85.20 \\
            L2 & 84.38 & 84.73 & 85.82 \\
            Geo & 85.54 & 85.60 & 85.74\\
        \end{tabular}
        \caption{Accuracy}
        \label{subtab:kd_acc}
    \end{subtable}
    \begin{subtable}{.3\linewidth}
        \centering
        \begin{tabular}{c|ccc}
            \diagbox{$\mathbf{f}_\perp$}{$\mathbf{f}_\parallel$} & \xmark & L2 & Geo \\
            \hline
            \xmark & 13.01 & 3.91 & 1.36 \\
            L2 & 2.51 & 2.26 & 0.97 \\
            Geo & 1.39 & 1.02 & 0.81 \\
        \end{tabular}
        \caption{Catastrophic Forgetting}\label{subtab:kd_for}
    \end{subtable}
    \begin{subtable}{.3\linewidth}
        \centering
        \begin{tabular}{c|ccc}
            \diagbox{$\mathbf{f}_\perp$}{$\mathbf{f}_\parallel$} & \xmark & L2 & Geo \\
            \hline
            \xmark & 17.60 & 4.73 & 1.16 \\
            L2 & 2.57 & 1.86 & 0.66 \\
            Geo & 1.32 & 0.80 & 0.36 \\
        \end{tabular}
        \caption{Zero-shot Degradation}\label{subtab:kd_deg}
    \end{subtable}
\end{table}

% \subsection{Subspace Rank}\label{subsec:subspace_rank}
% Here, we study the dimensionality of the subspaces learned during projector optimization.
% More specifically, we train our approach using a square projector $U \in \mathbb{R}^{d\times d}$ and examine the rank dimensionality via SVD decomposition and eigenvalue analysis.
% The results are reported in...
\section{Computational Requirements}
\label{sec:compute}

Subspace learning requires negligible compute overhead, and TASSO's training is faster than the main competitor (SnD): full training on $S^1$ (80k iters) requires $13.34$ TFLOPs for TASSO (both with and without subspace learning) and $15.49$ TFLOPs for SnD. Max VRAM use is $17.13$ GBs for all settings. Wall-clock time is $1h 36m 00s$ for TASSO (removing subspace learning is just 3s less, \ie, $1h 35m 57s$), while for SnD is $1h 50m 12s$.

%%%%%%%%%%%%%%%%%%%%%%%%%%%%%%%%%%%%%%
%\newpage
\section{Per-Domain Results}
\label{sec:confusion}
We report in this section the detailed results for each domain at each step of the incremental learning procedure attained during the Multidomain Task-Incremental Learning and the Multidomain Class-Incremental Learning experiments in the main document.
We present the $\mathbf{S^1}$ results on the MTIL benchmark in Tab.~\ref{tab:MTIL_matrix}, and the corresponding results on the MCIL benchmark in Tab.~\ref{tab:MCIL_matrix}. These tables corroborate the findings from the aggregate metrics: although performance in the more challenging MCIL setting is slightly lower than in the MTIL one, the gap is minimal, and the behavior across tasks and datasets remains largely consistent. 

\begin{table}[h]
\centering
\caption{Accuracy for each domain at each incremental training step on sequence $S^1$ in the MTIL benchmark.}
\resizebox{\textwidth}{!}{
\begin{tabular}{lcccccccc}
\toprule
 & \multicolumn{8}{c}{Dataset} \\
\cmidrule{2-9}
 & FGVC-Aircraft & DTD & EuroSAT & Flowers-102 & Food-101 & Oxford-Pets & Stanford-Cars & UCF-101 \\
\midrule
Original VLM & 23.91 & 44.39 & 42.22 & 67.40 & 83.69 & 87.27 & 65.51 & 64.26 \\
\midrule
FGVC-Aircraft & 53.74 & 44.56 & 42.74 & 66.95 & 83.58 & 86.92 & 65.32 & 64.10 \\
DTD & 53.98 & 80.14 & 45.65 & 66.83 & 83.39 & 87.30 & 64.97 & 64.21 \\
EuroSAT & 53.86 & 80.08 & 98.80 & 67.15 & 83.31 & 86.97 & 64.83 & 64.00 \\
Flowers-102 & 53.95 & 80.26 & 98.72 & 99.07 & 83.25 & 87.05 & 64.71 & 63.89 \\
Food-101 & 53.35 & 79.14 & 98.64 & 98.58 & 90.64 & 86.97 & 64.53 & 63.73 \\
Oxford-Pets & 53.20 & 78.90 & 98.64 & 98.54 & 90.54 & 95.45 & 64.31 & 63.86 \\
Stanford-Cars & 51.25 & 79.31 & 98.64 & 98.25 & 90.52 & 95.48 & 84.34 & 63.60 \\
UCF-101 & 50.83 & 78.96 & 98.60 & 98.34 & 90.37 & 95.48 & 84.09 & 89.24\\
\bottomrule
\end{tabular}}
\label{tab:MTIL_matrix}
\end{table}
%%%%%%%%%%%%%%%%%%%%%%%%%%%%%%%%%%%%%%
% \subsection{MCIL Benchmark}

%%%%%%%%%%%%%%%%%%%%%%%%%%%%%%%%%%%%%%
\begin{table}[h]
\centering
\caption{Accuracy for each domain at each incremental training step on sequence $S^1$ in the MCIL benchmark.}
\resizebox{\textwidth}{!}{
\begin{tabular}{lcccccccc}
\toprule
 & \multicolumn{8}{c}{Dataset} \\
\cmidrule{2-9}
 & FGVC-Aircraft & DTD & EuroSAT & Flowers-102 & Food-101 & Oxford-Pets & Stanford-Cars & UCF-101 \\
\midrule
Original VLM & 23.91 & 36.52 & 32.80 & 67.40 & 83.15 & 87.05 & 65.51 & 63.68 \\
\midrule
FGVC-Aircraft & 52.84 & 36.94 & 33.05 & 66.91 & 83.06 & 86.70 & 65.32 & 63.55 \\
DTD & 52.69 & 78.31 & 34.30 & 66.42 & 82.40 & 87.11 & 64.97 & 63.63 \\
EuroSAT & 52.51 & 78.07 & 98.31 & 66.71 & 82.32 & 86.84 & 64.83 & 63.49 \\
Flowers-102 & 52.72 & 77.96 & 98.17 & 98.94 & 82.26 & 86.86 & 64.71 & 63.34 \\
Food-101 & 52.27 & 76.00 & 98.26 & 98.42 & 90.22 & 86.75 & 64.53 & 63.28 \\
Oxford-Pets & 52.09 & 75.71 & 98.31 & 98.38 & 90.12 & 95.31 & 64.31 & 63.44 \\
Stanford-Cars & 50.17 & 76.00 & 98.40 & 98.09 & 90.01 & 95.34 & 84.34 & 63.15 \\
UCF-101 & 49.80 & 75.59 & 98.37 & 98.17 & 89.84 & 95.42 & 84.09 & 88.69\\
\bottomrule
\end{tabular}}
\label{tab:MCIL_matrix}
\end{table}
%%%%%%%%%%%%%%%%%%%%%%%%%%%%%%%%%%%%%%

%\hspace{.5em}

% More specifically, we use FGVC-Aircraft~\cite{maji2013fine} (containing $100$ airplane models), DTD~\cite{cimpoi2014describing} (containing $47$ patterns and textures), EuroSAT~\cite{helber2019eurosat} (containing $10$ types of satellite imagery), Flowers-102~\cite{nilsback2008automated} (containing $102$ species of flowers), Food-101~\cite{bossard2014food} (containing $101$ classes of foodstuffs), Oxford-Pets~\cite{parkhi2012cats} (containing $37$ species of cats and dogs), Stanford-Cars~\cite{krause20133d} (containing $196$ car models), and UCF-101~\cite{khurram2012dataset} (containing $101$ action recognition classes). 

\newpage

\section{Statistical Analysis}\label{sec:stats}
Here, we report a study on the stability of our approach across varying random seeds. The experimental results are shown in Tab. \ref{tab:stat}, where sequence $\mathbf{S^1}$ under the MTIL setting is executed multiple times with different random initializations. 
Looking at the results in the table, one can appreciate how the gains with respect to the closest competitor are preserved in all cases, falling well outside the confidence bounds of our measure. Moreover, the variance of the Forgetting and Degradation metrics is contained ($0.20$ and $0.18$, respectively) and is very close to that of the accuracy ($0.19$). As a final remark, we would like to highlight how the seed used in the main experimental results is the closest to the mean performance across seeds, with an average distance in all metrics of $0.08$.

\begin{table}[h]
\centering
%\captionsetup[subtable]{labelformat=simple}
%\renewcommand\thesubtable{}
%\begin{subtable}[t]{.45\linewidth}
    \centering
    \caption{%\small
    %\textbf{Table \thetable: }
    Statistical significance of TASSO on $\mathbf{S^1}$ of MTIL. For fairness, we used seed 1102 to match the one used by the main competitor SnD~\cite{yu2024select}.} %$\mu$ indicates mean, $\sigma$ indicates standard deviation.}
     \setlength{\tabcolsep}{12pt}
    \begin{tabular}{lccc}
    \toprule
    Seed & Accuracy & {Forgetting} & {Z.S. Degradation} \\
    \midrule
    $42$ & 85.73 & 0.89 & 0.45\\ % server 107
    $123$ & 86.03 & 0.65 & 0.74\\ % server 033
    $1234$ & 85.62 & 0.73 & 0.43\\ % server 086
    $12345$ & 85.39 & 1.21 & 0.79\\ % server 033
    $123456$ & 85.79 & 0.61 & 0.50\\ %server 086
    $1234567$ & 85.87 & 0.63 & 0.85\\ % server 004
    \hdashline
    $1102$ & 85.74 & 0.81 & 0.36 \\
    \midrule
    %$\mu\pm\sigma$ & 85.74\pms{0.19} & 0.79\pms{0.20} & 0.59\pms{0.18} \\
    Avg. & 85.74 & 0.79 & 0.59 \\
    Std. & 0.19 & 0.20 & 0.18 \\
    \bottomrule
    \end{tabular}
    \label{tab:stat}
\end{table}

 \newpage
 
\section{Details on Datasets and Sequences}\label{sec:datasets}
In this section, we report additional details on the dataset and task sequences used for TASSO evaluation. More specifically, Tab. \ref{tab:dsets} reports the number of classes and the amount of training and test samples for all datasets used, while Tab. \ref{tab:seqs} reports a detailed breakdown of the 8 sequences used in the evaluation.

\begin{table}[h]%{.45\linewidth}
    %\stepcounter{table}
    \centering
    \caption{ %\small
    %\textbf{Table \thetable: }
    Number of classes and samples in the training and test splits of the dataset used in TASSO. (*): Note that only a subset of the 1.28M training samples from ImageNet has been used.}
    \label{tab:dsets}
       \setlength{\tabcolsep}{12pt}
    \begin{tabular}{cccc}
    \toprule
        Dataset & Classes & Train & Test \\
        \hline
        FGVC-Aircraft~\cite{maji2013fine} & 100 & 3,334 & 3,333\\
        DTD~\cite{cimpoi2014describing} & 47 & 2,820 & 1,692\\ 
        EuroSAT~\cite{helber2019eurosat} & 10 & 13,500 & 8,100\\
        Flowers-102~\cite{nilsback2008automated} & 102 & 4,093 & 2,463 \\
        Food-101~\cite{bossard2014food} & 101 & 50,500 & 30,300\\
        Oxford-Pets~\cite{parkhi2012cats} & 37 & 2,944 & 3,669\\
        Stanford-Cars~\cite{krause20133d} & 196 & 6,509 & 8,041 \\
        UCF-101~\cite{khurram2012dataset} & 101 & 7,639 & 3,783\\
        \hline
        ImageNet-1k~\cite{deng2009imagenet} & 1000 & 100,000* & 50,000\\
        \bottomrule
    \end{tabular}
%\end{subtable}
\end{table}

\begin{table}[h]
    \centering
    \caption{Details on the 8 sequences used in the TASSO experimental evaluation.}
    \label{tab:seqs}
    \resizebox{\linewidth}{!}{\begin{tabular}{c|c:c:c:c:c:c:c:c}
        Sequence & $\mathcal{T}^1$ & $\mathcal{T}^2$ & $\mathcal{T}^3$ & $\mathcal{T}^4$ & $\mathcal{T}^5$ & $\mathcal{T}^6$ & $\mathcal{T}^7$ & $\mathcal{T}^8$\\
        \hline
        \multirow{2}{*}{$\mathbf{S^1}$} & FVGC & \multirow{2}{*}{DTD} & \multirow{2}{*}{EuroSAT} & Flowers & Food & Oxford & Stanford & UCF \\
		& Aircraft & & & 102 & 101 & Pets & Cars & 101 \\
        \hdashline
        \multirow{2}{*}{$\mathbf{S^2}$} & \multirow{2}{*}{DTD} & \multirow{2}{*}{EuroSAT} & Flowers & Food & Oxford & Stanford & UCF & FVGC\\
		& & & 102 & 101 & Pets & Cars & 101 & Aircraft \\
		\hdashline
		\multirow{2}{*}{$\mathbf{S^3}$} & \multirow{2}{*}{EuroSAT} & Flowers & Food & Oxford & Stanford & UCF & FVGC & \multirow{2}{*}{DTD}\\
		& & 102 & 101 & Pets & Cars & 101 & Aircraft & \\
        \hdashline
		\multirow{2}{*}{$\mathbf{S^4}$} & Flowers & Food & Oxford & Stanford & UCF & FVGC & \multirow{2}{*}{DTD} & \multirow{2}{*}{EuroSAT}\\
		& 102 & 101 & Pets & Cars & 101 & Aircraft & & \\
        \hdashline
		\multirow{2}{*}{$\mathbf{S^5}$} & Food & Oxford & Stanford & UCF & FVGC & \multirow{2}{*}{DTD} & \multirow{2}{*}{EuroSAT} & Flowers\\
		& 101 & Pets & Cars & 101 & Aircraft & & & 102 \\
        \hdashline
        \multirow{2}{*}{$\mathbf{S^6}$} & Oxford & Stanford & UCF & FVGC & \multirow{2}{*}{DTD} & \multirow{2}{*}{EuroSAT} & Flowers & Food\\
		& Pets & Cars & 101 & Aircraft & & & 102 & 101 \\
        \hdashline
        \multirow{2}{*}{$\mathbf{S^7}$} & Stanford & UCF & FVGC & \multirow{2}{*}{DTD} & \multirow{2}{*}{EuroSAT} & Flowers & Food & Oxford\\
		& Cars & 101 & Aircraft & & & 102 & 101 & Pets \\
        \hdashline
        \multirow{2}{*}{$\mathbf{S^8}$} & UCF & FVGC & \multirow{2}{*}{DTD} & \multirow{2}{*}{EuroSAT} & Flowers & Food & Oxford & Stanford\\
		& 101 & Aircraft & & & 102 & 101 & Pets & Cars \\
	\end{tabular}}
\end{table}

\section{Pseudocode of our Method}\label{sec:pseudocode}
Here, we report the pseudocode implementation of the TASSO algorithm for fine-tuning the vision encoder on a given task.
We also report the pseudocode of the evaluation pipeline, which includes the cross-task confusion matrix and the computation of aggregate metrics for reproducibility purposes.

\begin{algorithm}[h]
    \caption{TASSO Fine-tuning Protocol } %\fb[@chang ]{cleanup and update so the style matches the other algorithm - remove undefined symbols/notation (e.g. student...)}}
    \label{alg:fine-tune}
    \begin{algorithmic}[1]
        \REQUIRE reference dataset $\mathcal{X}^{\text{ref}}$, training steps $T$, hyperparameters $\alpha, \beta$
        \INPUT Current task data $\mathcal{T}^k = \{(\mathcal{D}^k, \mathcal{C}^k)\}$, previous step model $g^{k-1}=(g_i^{k-1}, g_t)$
        \OUTPUT Current step model $g^k=(g_i^k, g_t)$
        \STATE Initialization: $g_i^k \gets g_i^{k-1}$; freeze text encoder $g_t$
        \STATE Randomly initialize $A \in \mathbb{R}^{d\times r}$
        \FOR{$t=1, \dots, T$}
            \STATE $\mathcal{B}=\{(x,y)\}$ $\sim$ $\mathcal{D}^k$ \COMMENT{Sample minibatch from current task's data}
            \STATE $f_t(c)\gets g_t(c)$ for all $c\in\mathcal{C}^k$; \COMMENT{Text embeddings, $\|f_t(c)\|_2=1$}
            \STATE  $f_i^k(x)\gets g_i^k(x)$ for $x\in\mathcal{B}$;  \COMMENT{Current model image emeddings, $\|f_i^k(x)\|_2=1$}

            \COMMENT{Current task supervision (Cross-Entropy loss)}
            \STATE $\mathbf{s}(x) \gets \mathrm{softmax}([\langle f_i^k(x), f_t(c)\rangle \ \forall c\in\mathcal{C}^k])$
            \STATE $\mathcal{L}_{CE} \gets \frac{1}{|\mathcal{B}|}\sum_{(x,y)\in\mathcal{B}}\mathrm{CE}(\mathbf{s}(x),y)$

            \COMMENT{Orthonormal projector (QR reparameterization):}
            \STATE $[U_k,\_]\gets \mathrm{QR}(A)$ \COMMENT{decomposition in reduced mode, $U_k\in\mathbb{R}^{d\times r}$ and $U_k^\top U_k=I_r$}

            \COMMENT{Subspace learning:}
            \STATE $f_{\parallel}^k(x)\gets U_kU_k^\top f_i^k(x)$;\ \ $f_{\parallel}^t(c)\gets U_kU_k^\top f_t(c)$
            \STATE re-normalize $f_{\parallel}^k(x)$ and $f_{\parallel}^t(c)$ \COMMENT{$\|f_{\parallel}^k(x)\|_2=1, \|f_{\parallel}^t(c)\|_2=1$}
            \STATE $\mathbf{s}_{\parallel}(x)\gets \mathrm{softmax}([\langle f_{\parallel}^k(x),f_{\parallel}^t(c)\rangle\ \forall c\in\mathcal{C}^k])$

            \STATE $\mathcal{L}_{sub}\gets \frac{1}{|\mathcal{B}|}\sum_{(x,y)\in\mathcal{B}}\mathrm{CE}(\mathbf{s}_{\parallel}(x),y)$

            \STATE $\mathcal{B}^{\text{ref}}=\{x^{\text{ref}}\}$ $\sim$ $\mathcal{X}^{\text{ref}}$ \COMMENT{Sample minibatch from reference dataset}
            
            %\COMMENT{Teacher/Student embeddings on $\mathcal{B}^{ref}$:}
            \STATE $f_i^{k}(x^{\text{ref}})\gets g_i^k(x^{\text{ref}})$, \ \ $f_i^{k-1}(x^{\text{ref}})\gets g_i^{k-1}(x^{\text{ref}})$; \COMMENT{normalize both}

            \COMMENT{Decompose into \textit{task-specific} and \textit{task-irrelevant} subspaces:}
            \STATE $f_{\parallel}^{k}\gets U_kU_k^\top f_i^{k}$,\quad $f_{\perp}^{k}\gets f_i^{k}-f_{\parallel}^{k}$
            \STATE $f_{\parallel}^{k-1}\gets U_kU_k^\top f_i^{k-1}$,\quad $f_{\perp}^{k-1}\gets f_i^{k-1}-f_{\parallel}^{k-1}$

            % \COMMENT{Geodesic distance on the hypersphere:}
            % \STATE $\mathcal{L}_{KD}(a,b)\triangleq \cos^{-1}(\langle a,b\rangle)$

            \COMMENT{Geometry-aware KD:}
            \STATE $\mathcal{L}_{KD}\!\gets\!\frac{1}{|\mathcal{B}^{\text{ref}}|}\!\sum_{x^{\text{ref}}\in\mathcal{B}^{\text{ref}}}\!
            \left[\cos^{-\!1}(f_{\parallel}^{k}(x^{\text{ref}}),f_{\parallel}^{k\!-\!1}(x^{\text{ref}})) + \cos^{-\!1}(f_{\perp}^{k}(x^{\text{ref}}),f_{\perp}^{k\!-\!1}(x^{\text{ref}}))\right]$

            \COMMENT{Optimization objective:}
            \STATE $\mathcal{L}\gets \mathcal{L}_{CE} + \alpha\,\mathcal{L}_{\text{sub}} + \beta\,\mathcal{L}_{KD}$

            \STATE Update $(\theta_i^k, A)$; keep $g_t$ frozen \COMMENT{Update with AdamW}
        \ENDFOR
        \ENSURE $g^k=(g_i^k,g_t)$
    \end{algorithmic}
\end{algorithm}
% \subsection{Pseudocode for Continual Learning of the whole sequence}
% \subsection{Pseudocode for the evaluation}

% \begin{algorithm}[t]
%     \caption{TASSO Sequence Learning}
%     \label{alg:sequence}
%     \begin{algorithmic}[1]
%         \REQUIRE Text encoder $g_t$, vision encoder $g_i^0$
%         \INPUT Task sequence $\mathbf{S} = \{\mathcal{T}^1, \mathcal{T}^2,\dots,\mathcal{T}^8\}$
%         \OUTPUT Fine-tuned vision encoders $\{g_i^k,\; k=1,\dots,8\}$
%         \STATE $\mathcal{G}_i \gets \varnothing$ \COMMENT{Initialize the output set}
%         \STATE $g_i^{k-1} \gets g_i^0$ \COMMENT{Initialize previous-task encoder with the pretrained VLM} 
%         \FOR{$\mathcal{T} \in \mathbf{S}$}
%             \STATE $g_i^{k-1} \gets \text{TassoFineTune}(\mathcal{T}, g_i^{k-1})$ \COMMENT{Train encoder}
%             \STATE $\mathcal{G}_i \gets \mathcal{G}_i \cup \{g_i^{k-1}\}$ \COMMENT{Add new encoder to output set}
%         \ENDFOR
%         \ENSURE $\mathcal{G}_i = \{g_i^k,\; k=1,\dots,8\}$
%     \end{algorithmic}
% \end{algorithm}

% \fb[@chang ]{i commented the old algorithm, could you convert it to be similar to the one I added above from an old paper? P.S. you don't need to add boolean flags, you can simply use text to switch cases, e.g.: if MITL...}
% \clearpage
\begin{algorithm}[t]
    \caption{TASSO Evaluation Protocol}
    \label{alg:eval}
    \begin{algorithmic}[1]
        \REQUIRE Text encoder $g_t$, setting (MTIL/MCIL), number of tasks $K$, pretrained model accuracy on MTIL $\mathbf{p}_{\text{MTIL}} \in \mathbb{R}^K$ and MCIL $\mathbf{p}_{\text{MCIL}} \in \mathbb{R}^K$
        \INPUT Task Sequence $\mathbf{S} = \{\mathcal{T}^1, \mathcal{T}^2,\dots,\mathcal{T}^K\}$, Vision Encoders $\{g_i^k,\; k=1,\dots,K\}$
        \OUTPUT Mean Accuracy $\bar{a}$, Mean Forgetting $\bar{f}$, Mean Zero-Shot Degradation $\bar{d}$
        \IF{MCIL setting}
            \STATE $\mathcal{C}^k \gets \bigcup_{k'=1}^K \mathcal{C}^{k'} \;\; \forall k=1,\dots,K$ \COMMENT{Update the task class sets: under the MCIL setting, the class set is the union of all sets.}
            \STATE $\mathbf{b} \gets \mathbf{p}_{\text{MCIL}}$ \COMMENT{Select the appropriate upper-bound}
        \ELSE
            \STATE $\mathbf{b} \gets \mathbf{p}_{\text{MTIL}}$ \COMMENT{Select the appropriate upper-bound}
        \ENDIF
        \STATE $\mathbf{M} \gets \mathbf{0}_{K\times K}$ \COMMENT{Initialize an K-by-K matrix of zeros to store accuracy metrics}
        \FOR{$k_1 = 1,\dots,K$} \COMMENT{Training task loop - row}
            \FOR{$k_2 = 1,\dots,K$} \COMMENT{Evaluation task loop - column}
                \STATE $(\mathcal{D}, \mathcal{C}) \gets \mathcal{T}^{k_2} \in \mathbf{S}$ \COMMENT{Extract dataset and class set from the sequence.}
                \FOR{$(\mathbf{x}, y) \in \mathcal{D}$} \COMMENT{For each sample and label in the dataset}
                    \STATE $\hat{y} \gets \text{argmax}_{c\in\mathcal{C}} \left<g_i(\mathbf{x}), g_t(c)\right>$ \COMMENT{Compute the prediction by comparing the embedding of the image to all text embeddings}
                    \IF{$\hat{y} = y$}
                        \STATE $\mathbf{M}[k_1,k_2] \gets \mathbf{M}[k_1,k_2] + \frac{100}{|\mathcal{D}|}$ \COMMENT{Accumulate percent (top-1) accuracy}
                    \ENDIF
                \ENDFOR
            \ENDFOR
        \ENDFOR

        \COMMENT{Mean Accuracy}
        \STATE $\bar{a} \gets \frac{1}{K}\sum_{k_2=1}^K \mathbf{M}[K,k_2]$ \COMMENT{Average across tasks}
        
        \COMMENT{Mean Forgetting} %: diagonal minus worst later performance (exclude last task)}
        \STATE $\bar{f} \gets 0$
        \FOR{$k = 1,\dots,K-1$}
            \STATE $a \gets \mathbf{M}[k,k]$ \COMMENT{Fine-tuning accuracy}
            \STATE $\hat{a} \gets \min\limits_{k' = k+1,\dots,K} \mathbf{M}[k',k]$ \COMMENT{Worst accuracy on following incremental steps.}
            \STATE $\bar{f} \gets \bar{f} + (a - \hat{a})$
        \ENDFOR
        \STATE $\bar{f} \gets \frac{1}{K-1}\bar{f}$

        \COMMENT{Mean Zero-shot Degradation} %: baseline minus worst pre-training-stage performance (exclude first task)}
        \STATE $\bar{d} \gets 0$
        \FOR{$k = 2,\dots,K$}
            \STATE $\hat{a} \gets \min\limits_{k' = 1,\dots,k-1} \mathbf{M}[k',k]$ \COMMENT{Worst accuracy on unseen tasks.}
            \STATE $\bar{d} \gets \bar{d} + (\mathbf{b}[k] - \hat{a}$
        \ENDFOR
        \STATE $\bar{d} \gets \frac{1}{K-1}\bar{d}$

        \ENSURE $\bar{a}, \bar{f}, \bar{d}$
    \end{algorithmic}
\end{algorithm}

% \FloatBarrier
% \bibliography{main}

\end{document}